\documentclass[manuscript,screen]{acmart}
\usepackage{subcaption}
\usepackage{array}
\usepackage{makecell}

\usepackage{listings} 
\usepackage{xcolor}   
\usepackage{caption}  

\AtBeginDocument{%
  }

\setcopyright{acmlicensed}
\copyrightyear{2018}
\acmYear{2018}
\acmDOI{XXXXXXX.XXXXXXX}

\acmConference[Conference acronym 'XX]{Make sure to enter the correct
  conference title from your rights confirmation emai}{June 03--05,
  2018}{Woodstock, NY}
\acmISBN{978-1-4503-XXXX-X/18/06}

\begin{document}

\title{VLQA: The First Comprehensive, Large, and High-Quality Vietnamese Dataset for Legal Question Answering}

\author{Tan-Minh Nguyen}
\email{minhnt@jaist.ac.jp}
\orcid{0009-0001-3139-6349}
\affiliation{%
  \institution{Japan Advanced Institute of Science and Technology}
  \city{Ishikawa}
  \country{Japan}
}

\author{Hoang-Trung Nguyen}
\email{25025010@vnu.edu.vn}
\orcid{0009-0009-0758-441X}
\affiliation{%
  \institution{VNU University of Engineering and Technology}
  \city{Hanoi}
  \country{Vietnam}
}

\author{Trong-Khoi Dao}
\email{khoidt@vnu.edu.vn}
\affiliation{%
  \institution{VNU University of Law}
  \city{Hanoi}
  \country{Vietnam}}
  
\author{Xuan-Hieu Phan}
\email{hieupx@vnu.edu.vn}
\orcid{0000-0002-7640-9190}
\affiliation{%
  \institution{VNU University of Engineering and Technology}
  \city{Hanoi}
  \country{Vietnam}
}

\author{Ha-Thanh Nguyen}
\email{nguyenhathanh@nii.ac.jp}
\orcid{0000-0003-2794-7010}
\affiliation{%
 \institution{National Institute of Informatics}
 \city{Tokyo}
 \country{Japan}
 }

\author{Thi-Hai-Yen Vuong}
\authornote{Corresponding author.}
\email{yenvth@vnu.edu.vn}
\orcid{0000-0002-8019-7178}
\affiliation{%
  \institution{VNU University of Engineering and Technology}
  \city{Hanoi}
  \country{Vietnam}
}

\renewcommand{\shortauthors}{Tan-Minh Nguyen et al.}

\begin{abstract}
The advent of large language models (LLMs) has led to significant achievements in various domains, including legal text processing. Leveraging LLMs for legal tasks is a natural evolution and an increasingly compelling choice. However, their capabilities are often portrayed as greater than they truly are. Despite the progress, we are still far from the ultimate goal of fully automating legal tasks using artificial intelligence (AI) and natural language processing (NLP). Moreover, legal systems are deeply domain-specific and exhibit substantial variation across different countries and languages. The need for building legal text processing applications for different natural languages is, therefore, large and urgent. However, there is a big challenge for legal NLP in low-resource languages such as Vietnamese due to the scarcity of resources and annotated data. The need for labeled legal corpora for supervised training, validation, and supervised fine-tuning is critical. In this paper, we introduce the VLQA dataset, a comprehensive and high-quality resource tailored for the Vietnamese legal domain. We also conduct a comprehensive statistical analysis of the dataset and evaluate its effectiveness through experiments with state-of-the-art models on legal information retrieval and question-answering tasks.

\end{abstract}




\received{20 February 2007}
\received[revised]{12 March 2009}
\received[accepted]{5 June 2009}

\maketitle

\section{Introduction}
The legal system regulates a broad spectrum of everyday human activities such as finance, taxation, employment, marriage, etc., thereby exerting influence on both legal practitioners and the general population alike \cite{ponce2019global}. In recent years, advances in AI, particularly in NLP, have opened new avenues for enhancing access to legal information. The growing digitization of legal texts, combined with powerful language models, offers a compelling foundation for reducing the gap between legal knowledge and lay understanding \cite{zhong-etal-2020-nlp}. For instance, legal text summarization \cite{kanapala2019text,anand2022effective} and automated contract analysis \cite{hendrycks2021cuad,lippi2019claudette} can aid non-experts in navigating complex and often opaque legal language. Similarly, legal information retrieval (LIR) \cite{nguyen2024attentive,sugathadasa2019legal,louis-spanakis-2022-statutory} streamlines the process for practitioners to identify relevant statutory or case law, reducing the reliance on time-intensive manual efforts. Legal question answering (LQA) systems \cite{10.1609/aaai.v38i20.30232,Zhong_Xiao_Tu_Zhang_Liu_Sun_2020,10.1145/3594536.3595159} further hold promise for assisting individuals in obtaining preliminary legal insights without direct consultation.
However, existing research on LQA and LAR remains limited in scope, often focusing on narrow domains such as tax law or civil rights, and producing restricted forms of output like binary (yes/no) answers or short phrase spans.


To support the development of practical and scalable LQA systems, we introduce VLQA, the first comprehensive, large-scale, and expert-annotated dataset for the Vietnamese legal domain. VLQA comprises over 3,000 real-world legal questions raised by Vietnamese citizens, each meticulously annotated by legal professionals with references to relevant statutory articles drawn from a corpus of approximately 59,000 legal provisions, as summarized in Table~\ref{tab:compare_lqa}. To the best of our knowledge, VLQA represents the largest real-world expert-verified LQA dataset covering any statutory domain. VLQA offers four key contributions. 
First, the questions are sourced from public legal consultation platforms, ensuring that the dataset reflects genuine legal concerns encountered by everyday citizens. 
Second, the underlying legal corpus spans the entirety of Vietnamese statutory law, presenting a realistic and challenging information retrieval scenario for evaluating LIR and LQA systems. 
Third, the annotation process is conducted by senior law students under the supervision of an experienced legal expert. This process involves validating and refining the relevant legal articles and corresponding answers to ensure the overall quality, accuracy, and reliability of the dataset.
Finally, VLQA provides both relevant articles and long-form answers, addressing two fundamental LIR and LQA tasks. By making the dataset publicly available, we aim to advance research in legal NLP and promote the development of accessible, trustworthy legal assistance tools that help bridge the gap between legal systems and the public.

We establish strong baselines on the VLQA dataset by evaluating various state-of-the-art retrieval and LQA methods. Our retrieval experiments cover both traditional lexical models (e.g., BM25, fastText) and dense neural retrieval approaches (i.e., BERT, RoBERTa), offering valuable insights into their relative effectiveness in the statutory legal domain. For LQA, we benchmark extractive models based on transformer encoders (such as BERT and RoBERTa), generative models (e.g., BARTPho and ViT5), and recent LLMs, including open-weight systems (Qwen2.5, Llama 3.1, DeepSeek-V3) and commercial offerings (GPT-4o, GPT-4o-mini).
The performance of these systems is assessed through both automated metrics and manual qualitative analysis. While the LLMs demonstrate high alignment with gold answers under automated evaluation, human evaluation reveals that their outputs, despite being fluent and well-structured, might contain factual inaccuracies or hallucinated elements. These findings highlight a significant disparity between superficial language generation and robust legal reasoning, indicating opportunities for advancement in both evaluative frameworks and methodological approaches. The dataset and source code will be publicly released soon. In summary, our main contributions are:
\begin{itemize}
    \item We introduce the first comprehensive, large-scale, and expert-annotated Vietnamese benchmark for LQA, comprising 3,129 legal questions posed by laypeople, meticulously verified and labeled by professionals with detailed answers and citations to relevant statutory provisions, drawn from an expansive knowledge corpus of 59,636 law articles. 
    \item We conduct a thorough evaluation of state-of-the-art LIR and LQA methods via both automated metrics and human evaluation. Our analysis reveals that, while LLMs can generate fluent and detailed responses, they might include incorrect or hallucinated elements, which can ultimately lead to unreliable conclusions. Such errors are often subtle and difficult for laypersons without legal expertise to identify, underscoring the current limitations of LLMs in supporting juridical decision-making tasks.
    \item We will publicly release the dataset, source code, and checkpoints to facilitate reproducibility and support research of information retrieval and question answering in juridical scenarios.
\end{itemize}

\begin{table}[t]
\caption{Summary of statistics of existing legal question-answering datasets. To the best of our knowledge, VLQA is the largest expert-annotated real-world LQA dataset that covers any statutory domains.}
\label{tab:compare_lqa}
\resizebox{\linewidth}{!}{
\begin{tabular}{l|ccc|c|c|c|c|c|c}
\hline
\textbf{Dataset} & \multicolumn{3}{c|}{\textbf{Average \# of words}} & \textbf{\# Ques.} & \textbf{\# Evid.} & \textbf{Answer type}    & \textbf{Domain} & \textbf{Source} & \textbf{Lang.} \\
                 & Ques.          & Evid.           & Ans.           &                   &                   &                         &                 &                 &                \\ \hline
COLIEE 2024 \cite{10.1007/978-981-97-3076-6_8}     &      40.24          &     73.09            & 1              & 1,206             & 768               & Binary                  & Civil law       & Law exam        & ja, en         \\
JEC-QA \cite{Zhong_Xiao_Tu_Zhang_Liu_Sun_2020}          & 47             & 58              & 15             & 26,365            & 3382              & MCQ            & Statutory law   & Law exam        & zh             \\
PrivacyQA \cite{ravichander-etal-2019-question}       & 8              & 3,237           & 140            & 1,750             & 335                & Multi-span              & Privacy policy  & Jurists         & en             \\
EQUALS \cite{10.1145/3594536.3595159}          & 32             & 252             & 69             & 6,914             & 3,081             & Long-form               & Statutory law   & Web forum       & zh             \\
LLeQA \cite{10.1609/aaai.v38i20.30232}           & 15             & 1,857           & 264            & 1,868             & 27,942            & Long-form               & Statutory law   & Jurists         & fr             \\
ALQAC 2023 \cite{10299527}      & 24.91          & 184.07          & 1.94           & 100               & 2,131             & Binary, MCQ & Statutory law   & Jurists         & vi             \\ \hline
VLQA (ours)      & 19.78          &     303.49           & 216.74         & 3,129             & 59,636            & Long-form      & Statutory law   & Web forum       & vi             \\ \hline 
\end{tabular}
}
\end{table}

\begin{table}[t]
\caption{A data sample in VLQA, which supports fundamental tasks in legal NLP: document retrieval and question answering. All examples in this study are translated from Vietnamese for illustration.}
\begin{tabular}{l|p{12cm}}
\hline
Question          & Do taxpayers conducting electronic transactions in the tax field have to use other transaction methods?                 \\
Answer            & Taxpayers who have conducted electronic transactions in the tax field do not have to perform other transaction methods. \\
Relevant articles & Law\_id: 38/2019/QH14, Article\_id: 8                                                                              \\
                  & Law\_id: 38/2019/QH14, Article\_id: 16                                                                             \\ \hline
\end{tabular}
\end{table}

The remainder of this study is organized as follows. Section~\ref{sec:related_works} reviews existing legal NLP benchmarks across various domains and tasks. Section~\ref{section:dataset} describes the construction and key characteristics of the VLQA dataset. Section~\ref{sec:models} outlines the design and implementation of baseline models. Section~\ref{sec:experiment} explains the experimental setup and evaluation metrics. Section~\ref{sec:results} presents the performance results of different models on VLQA and reveals in-depth analyses. Finally, Section~\ref{sec:conclusion} concludes the paper and discusses promising directions for future work.

\section{Related works}
\label{sec:related_works}
To support the development of reliable models for various legal NLP tasks, the research community has introduced several benchmark datasets in recent years. 
For example, \citet{duan2019cjrc} proposed CJRC, a Chinese judicial reading comprehension dataset comprising over 50,000 pairs of (question, answer) generated from the fact descriptions of 10,000 legal cases. 
\citet{xiao2018cail2018} presented CAIL2018, the first large-scale dataset for legal judgment prediction (LJP) in Chinese, comprising more than 2.6 million criminal cases. These cases were sourced from the official website of the Supreme People’s Court of China and are associated with approximately 400 legislative articles and criminal charges.
Similarly, \citet{medvedeva2018judicial} developed a legal judgment prediction dataset consisting of 11,532 admissible cases from the European Court of Human Rights (ECtHR) as of September 2017. Meanwhile, \citet{semo-etal-2022-classactionprediction} introduced the first class action LJP dataset, comprising 5,459 lost cases and 5,290 won cases. \citet{xiao2019cail2019} released CAIL2019-SCM, containing approximately 9,000 triplets of cases for detecting similarity between cases. Additionally, \cite{de2018rulingbr} proposed a text summarization dataset of legal texts in Portuguese. contains about 10,000 rulings from the Brazilian Federal Supreme Court. Likewise, \citet{manor-li-2019-plain} introduced the benchmark of summarization of legal contracts in plain English, including 446 sets of contract actions and corresponding reference summaries. 

Statutory article retrieval remains an underexplored area in legal NLP due to the scarcity of large, high-quality datasets. Existing resources include the COLIEE Statute Law corpus \cite{10.1007/978-981-97-3076-6_8}, BSARD \cite{louis-spanakis-2022-statutory}, and ALQAC \cite{10299527}. The COLIEE dataset comprises 1,206 questions from the Japanese bar examination, annotated with references to articles in the Japanese Civil Code. Similarly, ALQAC consists of 100 question–article pairs curated by jurists and legal experts. Both datasets primarily target bar exam question-answering, which differs from real-world legal queries posed by non-experts. Bar exam questions are typically domain-specific, technical, and designed for legal professionals or students, while everyday legal inquiries from citizens are more general and straightforward. Recently, \citet{louis-spanakis-2022-statutory} introduced BSARD, a statutory article retrieval dataset in France with 1,108 legal questions and annotations referring to 22,633 articles, collected from legal consultant organizations. Similarly, VLQA also sources its questions from actual citizen inquiries. However, VLQA distinguishes itself by its broader scale and coverage, encompassing 3,129 annotated questions and a comprehensive corpus of nearly 60,000 Vietnamese statutory articles, offering a foundation for building practical, data-driven legal information retrieval systems.

Addressing LQA has long presented a significant challenge in legal NLP due to the intricate nature of legal language, rigid document structures, and the necessity for nuanced reasoning across temporal and logical dimensions. To advance this field, a variety of datasets have been developed, each reflecting different subdomains and tasks. For example, \citet{Zhong_Xiao_Tu_Zhang_Liu_Sun_2020} introduced a large-scale multiple-choice QA dataset from China’s National Judicial Examination, comprising 26,365 questions designed to evaluate the legal proficiency of professionals. Similarly, the COLIEE competition \cite{10.1007/978-981-97-3076-6_8} focuses on binary (``Yes''/``No'') answers to bar exam questions in Japanese civil law, emphasizing short-form, high-precision reasoning.
In contrast, long-form LQA has garnered growing attention, where systems must generate complete, comprehensive answers. PrivacyQA \cite{ravichander-etal-2019-question}, for instance, includes 1,750 questions over 335 privacy policy documents, with answer spans averaging 140 words. ViRHE4QA, a Vietnamese dataset, targets university training regulations and contains 9,758 annotator-generated questions across 294 legal articles. Most recently, LLeQA \cite{10.1609/aaai.v38i20.30232} introduced a long-form dataset from Belgian legal consulting platforms, comprising 1,868 real-world questions and answers covering diverse statutory domains, with average question and answer lengths of 15 and 264 words, respectively.
Following this trajectory, VLQA presents a great leap forward. It comprises 3,129 high-quality legal questions posed by Vietnamese citizens, annotated and verified by legal professionals. Unlike prior Vietnamese datasets, which focus on narrow domains or professional questions, VLQA is grounded in real-world, multi-domain inquiries. We envision VLQA as a foundational resource for advancing reliable and practically applicable legal question-answering systems, thereby contributing to narrowing the gap between legal knowledge and public accessibility.

\section{The VLQA Dataset}
\label{section:dataset}
This section presents the Vietnamese Legal Question Answering (VLQA) dataset, comprising expert-annotated 3,129 triplets of \{question, articles, answer\} and 59,636 articles across 27 domains in Vietnamese Laws, which address two fundamental legal NLP tasks: \textbf{legal article retrieval} and \textbf{legal question-answering}. We also provide comprehensive analyses in various aspects to offer statistical insights into the VLQA dataset. 

\subsection{Dataset Construction}
The dataset is constructed through three meticulous steps. First, we construct an expansive article-based legal corpus based on Vietnamese legislation. Then we collect and refine legal questions and their answers from legal consultation forums that feature user-generated concerns. Data triplets are validated and annotated by law experts to improve the accuracy and professionalism of the answers. Finally, VLQA is divided into three subsets: training, development, and testing. 

\begin{figure*}[ht]
    \centering
    \includegraphics[width=1\linewidth]{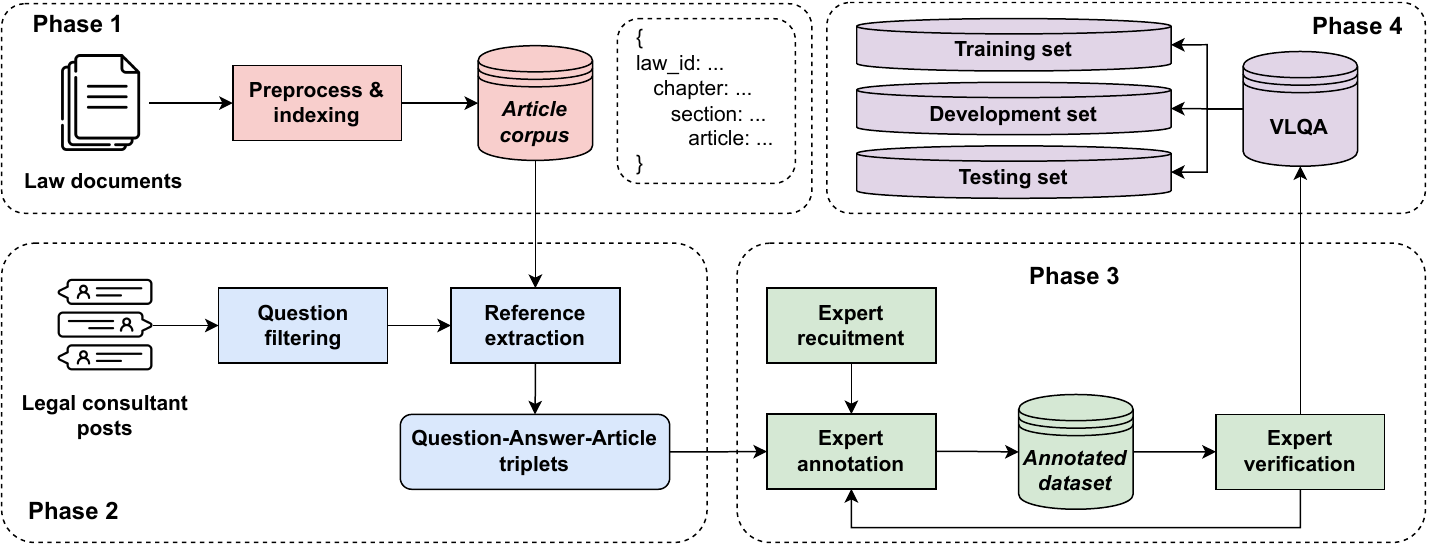}
    \caption{The detailed process of VLQA construction. This process involves four phases: article corpus collection (phase 1), question-answer-article triplets collection (phase 2), expert verification (phase 3), and data splitting (phase 4).}
    \label{fig:dataset-construction}
    \Description[Dataset Construction]{An illustration of the dataset construction process. All examples in this study are translated from Vietnamese for illustration.}
\end{figure*}

\subsubsection{Expansive article corpus collection} 

The Vietnamese legal system exhibits all the key characteristics of a civil law jurisdiction, including the use of legal documents that comprehensively regulate specific areas of law, such as civil rights or education. These documents serve as valuable resources for collecting laws across a wide range of subjects. In our study, we consider 2,162 legal documents spanning 27 common domains within the Vietnamese legal framework, as detailed in Table \ref{tab:state_legaldoc}. In addition to extracting individual legal articles, we also capture the structural hierarchy of each document (i.e., document, part, section, chapter), which helps provide context and an overview of article content. During the collection process, we exclude documents with identical titles but different subjects to avoid ambiguity.
During pre-processing, we apply regular expressions to correct typographical errors and remove misplaced characters (e.g, newlines, tabs). We also remove articles that were repealed by past laws but are still present in the source documents. Finally, we construct a legal corpus consisting of 59,636 articles, organized according to a hierarchical structure. These articles serve as the basic unit for the retrieval task. 

\subsubsection{Question-answer-article triplets collection} 
With the growing accessibility of online legal consultation platforms, an increasing number of individuals are turning to these services for professional advice on legal matters. In this study, we collect approximately 430,000 raw legal questions posted by laypersons across various domains, including marriage, taxation, and employment, sourced from well-known legal advice websites in Vietnam. 
During the question collection process, we identify and remove questions that reference entire legal documents, repealed articles, or refer to articles not included in the corpus. We further group duplicated questions that share the same content but differ in categorization tags. 
The question-answer pairs contain explicit textual mentions of legal provisions (e.g., ``Article 5 of the Civil Law 2015'', ``Article 2 of this code''). We employ regular expression-based parsing to extract these references and align them with their corresponding articles in the legal corpus. As the collected answers are human-generated, there is a lack of a unified format in these references. Hence, we filter out questions that cannot extract relevant law articles. 
Responses on legal consultation forums follow a recognizable structure, typically beginning with an analysis of the question, followed by references to relevant legal articles, and concluding with a summarized answer. To exploit these structured responses, we employ regular expressions to identify and extract relevant segments while filtering out irrelevant content.
To improve the consistency of the question-answer pairs, we use regular expressions and the SpaCy\footnote{\url{https://spacy.io/universe/project/spacy_fastlang}} language detection package to filter out informal phrases, non-Vietnamese words, icons, and irrelevant characters from the text. Misspelled and typographical errors are corrected using regular expressions.

\subsubsection{Expert validation}


Data annotation plays a critical role in improving the quality and readability of the dataset. Thus, we employ a professional annotation team with five senior law students as annotators and an experienced legal expert as a reviewer and supervisor. 
The data annotation process is carried out by law students, who are responsible for validating and refining the collected answers as needed. Before the official annotation phase, they receive training and detailed guidance from a legal reviewer to ensure consistency and accuracy. The final stage of quality assurance is conducted by a legal expert, who reviews the annotated data, performs thorough proofreading, and rectifies any remaining errors.

\begin{figure}
    \centering
    \includegraphics[width=\linewidth]{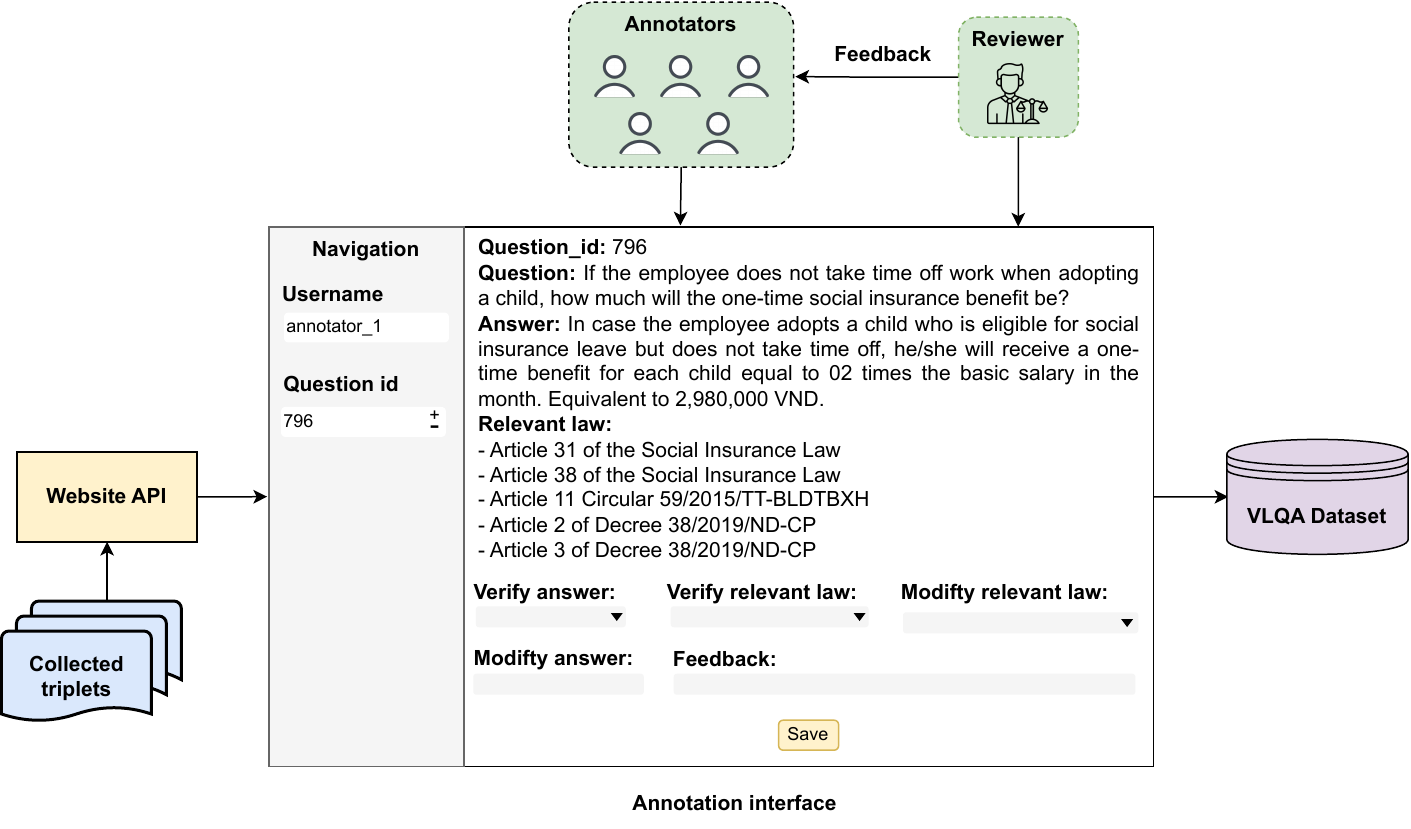}
    \caption{The interface of the data annotation framework for VLQA.}
    \label{fig:annotation_interface}
\end{figure}

We develop a user-friendly and systematic annotation website for annotators using Streamlit API\footnote{https://streamlit.io/}, as presented in Figure \ref{fig:annotation_interface}. Annotators are provided with the question, answer, and associated relevant law. They are required to verify whether the current answer and relevant law are correct and complete. Each question-answer pair is annotated independently by two annotators to reduce bias from an individual. 
Annotators are instructed to provide the following information for each sample:
\begin{itemize}
    \item Verify answer: Label whether the answer is True or False.
    \item Modify answer: Provide feedback on the answer. The annotator should provide the correct answer and specify whether the raw answer is incomplete, excessive, etc.
    \item Verify relevant law: Label whether the relevant laws are True or False.
    \item Modify relevant law: This annotator should select the correct relevant laws.
    \item Feedback: The annotator should provide a detailed explanation or comment about the dataset if necessary. We gather comments from annotators and the reviewer for further data processing.
\end{itemize}

\begin{table}[]
\caption{Statistics of reviewer-verified data.} 
\label{tab:diff_anno}
\begin{tabular}{l|cc}
\hline
                                             & \textbf{Count} & \textbf{Percentage} \\ \hline
Exact match (answer and articles unchanged)  & 2321           & 74.18\%             \\
Expert-modified articles, answer unchanged   & 464            & 14.83\%             \\
Expert-modified answer, articles unchanged   & 119            & 3.8\%               \\
Both answer and articles modified by reviewer & 225            & 7.19\%              \\ \hline
\multicolumn{1}{c|}{Total}          & 3,129          & 100\%               \\ \hline
\end{tabular}
\end{table}

Once the annotation task is completed, the quality of the annotated data is evaluated by the expert reviewer, as presented in Table \ref{tab:diff_anno}. During the review process, the reviewer will assess the data according to three standard aspects. 
\begin{itemize}
    \item Clearness: Whether the question is specific and well-formed to retrieve relevant laws.
    \item Validity: Whether the provided answer cites valid and current statutory articles. Any sample referencing repealed or outdated legal documents is discarded.
    \item Fluency: Whether the answer is coherent and interpretable for laypersons, with grammatically correct and logically connected sentences.
\end{itemize}

Data that does not meet the quality requirements will be revised or removed following the reviewer's instructions. Table \ref{tab:diff_anno} presents a detailed analysis of annotated outputs compared to reviewer-verified annotations for question answering and article retrieval. A majority of samples (74.18\%) demonstrate full agreement, where both the annotated answer and relevant legal articles align with reviewer judgments. In 14.83\% of cases, the legal reviewer retained the answer but revised the supporting articles, implying that while the answers were valid, alternative legal justifications were more appropriate or precise. Conversely, 3.80\% of instances saw the legal basis unchanged, but the answer needs correction, highlighting edge cases where subtle legal reasoning may have led to inaccurate decisions. Notably, 7.19\% of samples required changes to both the answer and supporting articles, as seen in Table \ref{tab:modify_both_answer_article}. In this context, Decree 38/2019/ND-CP has been amended by the most recent legislative update, Decree 24/2023/ND-CP, which took effect on July 1, 2023. As a result, the original answer of ``1,490,000 VND/month'' has been revised to ``1,800,000 VND/month'' to reflect the updated legal provision.

\begin{table}[t]
\caption{Example of both answer and article modified by the expert during the data review process, where repealed law articles lead to an incorrect answer.}
\label{tab:modify_both_answer_article}
\begin{tabular}{lp{5cm}p{5cm}}
\hline
                           & \multicolumn{1}{c}{\textbf{Original}}              & \multicolumn{1}{c}{\textbf{Verified}}                       \\ \hline
\textbf{Answer}            & Currently, the basic salary is \textit{1,490,000 VND/month}. & From July 1, 2023, the basic salary is \textit{1,800,000 VND/month}. \\
\textbf{Relevant articles} & Article 2 of \textit{Decree 38/2019/ND-CP}                  & Article 3 of \textit{Decree 24/2023/ND-CP}                           \\
\textbf{Context }          & Article 2. Subjects of application                 & Article 3. Basic salary                                     \\ \hline
\end{tabular}
\end{table}

\subsubsection{Data splitting}
\begin{figure}
    \centering
    \includegraphics[width=0.65\linewidth]{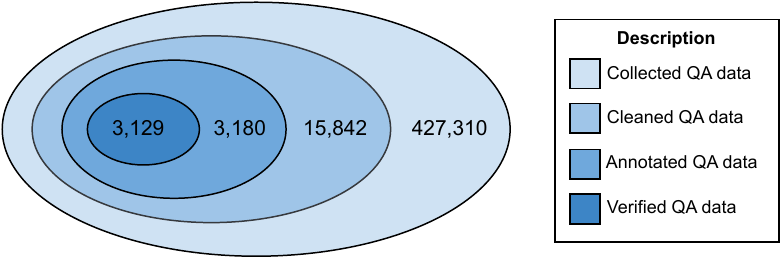}
    \caption{The number of samples in each processing step.}
    \label{fig:QA_sets}
\end{figure}

Finally, we construct an expert-annotated real-world legal QA dataset with 3,129 triplets of \{question, articles, answer\}, as shown in Figure \ref{fig:QA_sets}. The verified QA data is then partitioned into three subsets: training, validation, and testing, with a ratio of 7:1:2.

\subsection{Dataset Analysis}

\subsubsection{Overall Statistic}
\begin{figure*}[!tbp]
    \centering
    \begin{subfigure}[b]{0.47\textwidth}
        \centering
        \includegraphics[width=\linewidth]{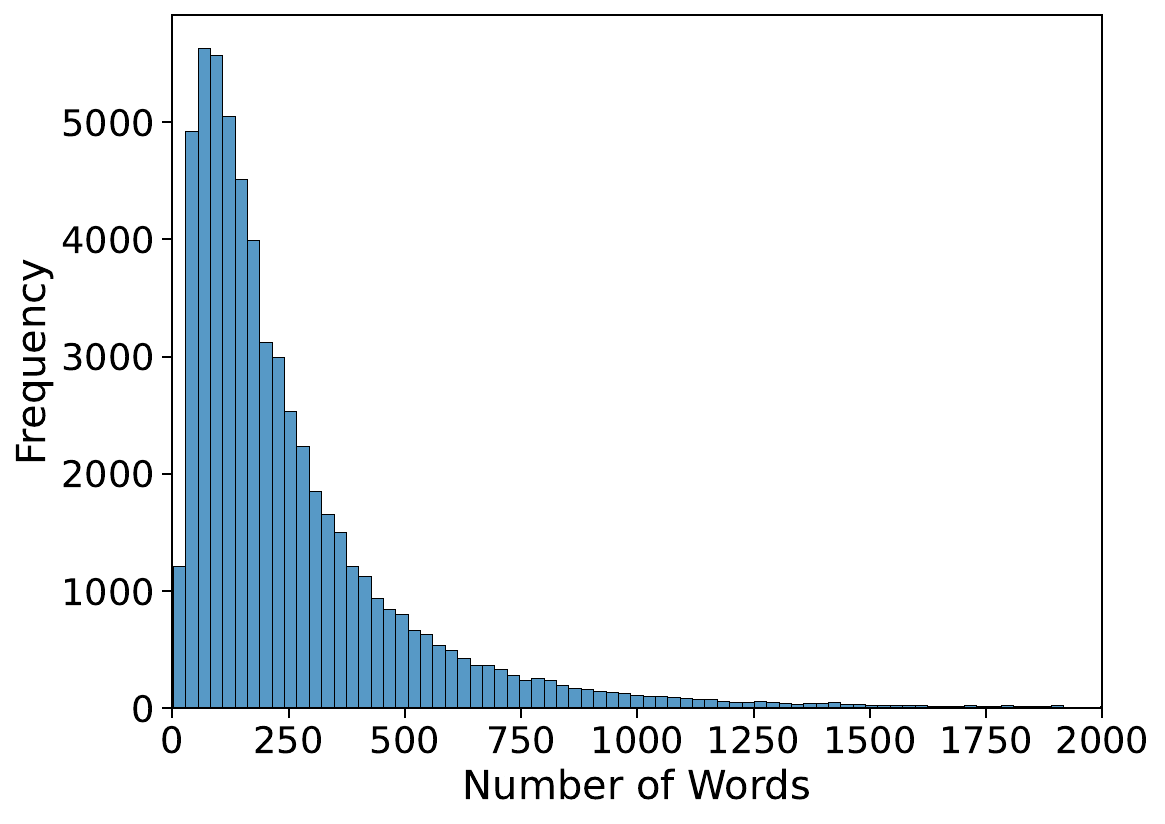}
        \caption{Article length.}
        \label{fig:article_length}
    \end{subfigure}
    \hfill
    \begin{subfigure}[b]{0.45\textwidth}
        \centering
        \includegraphics[width=\linewidth]{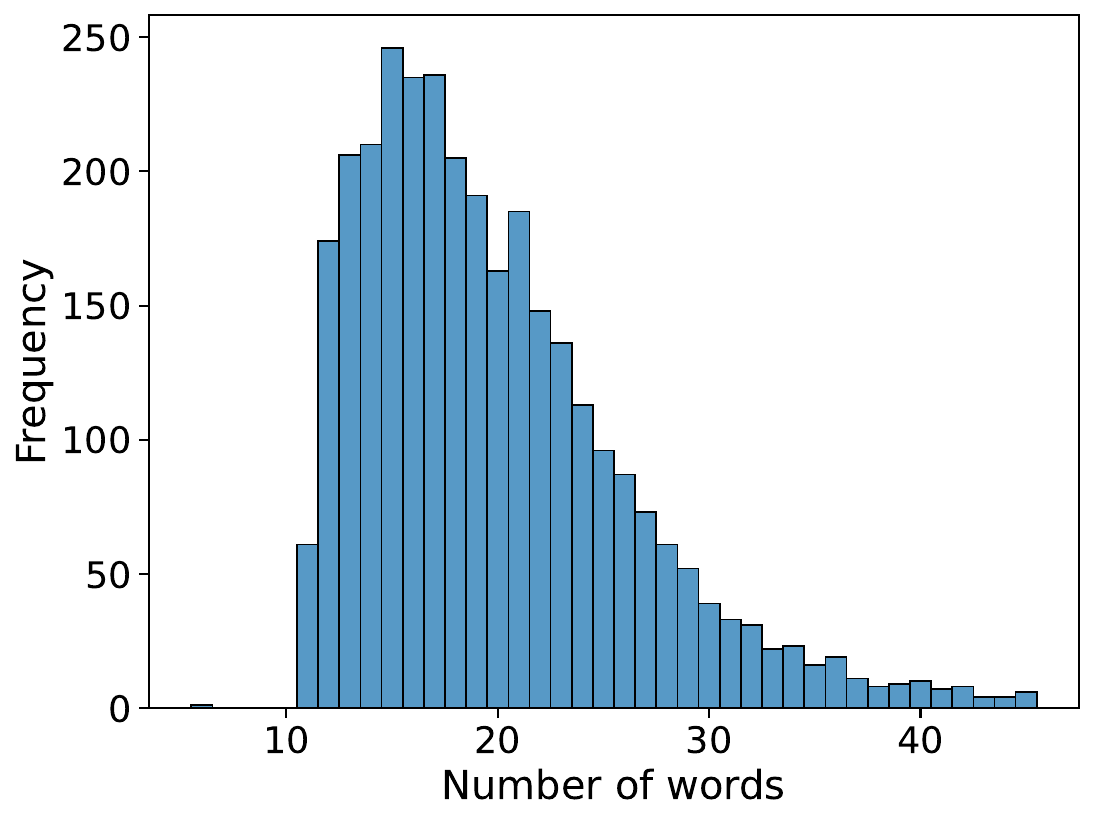}
        \caption{Question length.}
        \label{fig:qa_distribution_ques}
    \end{subfigure}
    \vfill
    \begin{subfigure}[b]{0.45\textwidth}
        \centering
        \includegraphics[width=\linewidth]{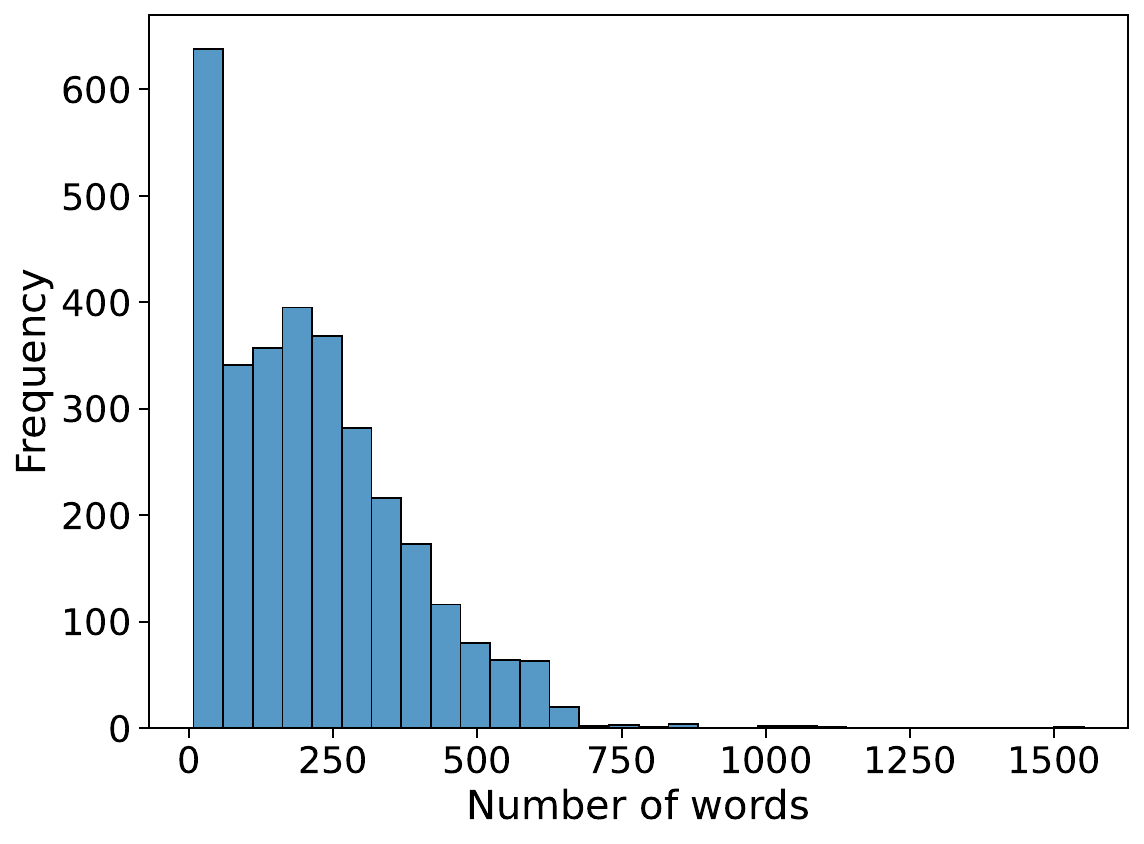  }
        \caption{Answer length.}
        \label{fig:qa_distribution_answer}
    \end{subfigure}
    \hfill
    \begin{subfigure}[b]{0.45\textwidth}
        \centering
        \includegraphics[width=\linewidth]{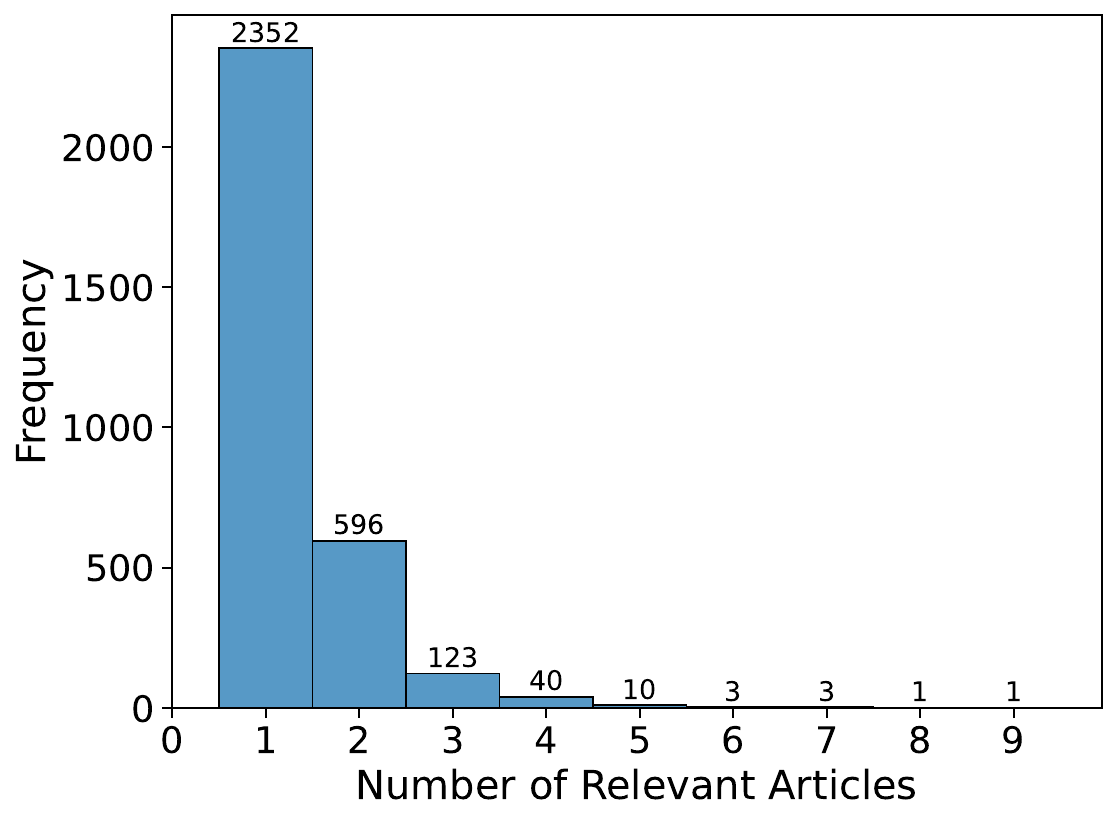}
        \caption{Number of relevant articles per question.}
        \label{fig:no_rel_law}
    \end{subfigure}
    \caption{Data statistics of VLQA.}
    \label{fig:len_distribution_vlqa}
    \Description[4 distributions]{Length distribution of article, question, and answer in VLQA and the number of relevant articles per question.}
\end{figure*}

Table \ref{tab:compare_lqa} reports a comprehensive comparison of VLQA and other LQA benchmarks on several factors: number of words per textual input, number of questions, number of evidences (law articles), answer type, covered domain, data source, and used languages. To the best of our knowledge, VLQA is the largest expert-annotated real-world LQA dataset, with long-form answers verified by experts that cover any statutory domains. Its distinct questions mirror real-world scenarios when laypeople seek legal consultants for advice. The significant size of the article corpus also poses challenges in locating and extracting relevant information within a longer context, supporting the development of strong and reliable legal AI systems. Figure \ref{fig:len_distribution_vlqa} shows the length distribution of questions, answers, and articles, along with the number of relevant articles per question, with a typical left-skewed pattern. Notably, the maximum and average number of relevant articles per question are 9 and 1.33, respectively, highlighting the complexity of VLQA.  



\begin{table}[t]
    \caption{Statistical analysis of law articles in the VLQA.}
    \label{tab:state_legaldoc}
    \centering
    \resizebox{\columnwidth}{!}{%
    \begin{tabular}{l|c|c|c}
    \hline  \textbf{Document Topic} & \textbf{Number of Chapters} & \textbf{Number of Sections} & \textbf{Number of Articles} \\
    \hline Currency and Banking 
     & 366 & 494 & 2475 \\
     Labor and Wages
    & 407 & 557 & 2440 \\
     Business
    & 487 & 656 & 3816 \\
     Investment
    & 260 & 383 & 2017 \\
     Commerce
    & 651 & 937 & 4748 \\
     Administrative Violations
    & 227 & 372 & 2372 \\
     Administrative System
    & 1381 & 1784 & 8566 \\
     Urban Development
    & 137 & 220 & 1122 \\
     Accounting and Auditing
    & 68 & 120 & 658 \\
     Public Finance
    & 310 & 404 & 1966 \\
     Taxes, Fees, and Charges 
    & 218 & 260 & 1278 \\
     Real Estate
    & 213 & 326 & 1701 \\
     Import and Export
    & 124 & 181 & 870 \\
     Information Technology
    & 318 & 418 & 2012 \\
     Civil Rights
    & 218 & 364 & 2218 \\
     Intellectual Property
    & 73 & 102 & 737 \\
     Sports and Healthcare
    & 406 & 568 & 2494 \\
     Insurance
    & 193 & 268 & 1432 \\
     Transportation and Logistics
    & 417 & 555 & 3017 \\
     Criminal Liability
    & 207 & 244 & 2197 \\
     Litigation Procedure
    & 230 & 300 & 2213 \\
     Legal Services
    & 102 & 117 & 642 \\
     Education
    & 356 & 419 & 1896 \\
     Securities
    & 67 & 135 & 799 \\
     Culture and Society
    & 255 & 331 & 1530 \\
     Natural Resources \& Environment
    & 404 & 581 & 2922 \\
     Other fields
    & 221 & 313 & 1498 \\
    \hline  \textbf{Total} & 8,316 & 11,409 & 59,636 \\
    \hline
    \end{tabular}
    }
\end{table}

The dataset summarized in Table \ref{tab:state_legaldoc} comprises a wide-ranging collection of legal articles that capture the structural and thematic diversity of the Vietnamese legal system. The Administrative System domain is the most extensive, encompassing 8,566 articles distributed across 1,381 chapters, reflecting its broad regulatory coverage and institutional complexity. In contrast, specialized domains such as Intellectual Property and Securities contain fewer articles, 737 and 799, suggesting a narrower legislative focus and more concentrated regulatory frameworks. This variation highlights the heterogeneity in legislative scope across different legal sectors.

\subsubsection{Question Diversity}
\begin{figure}[t]
    \centering
    \includegraphics[width=0.75\linewidth]{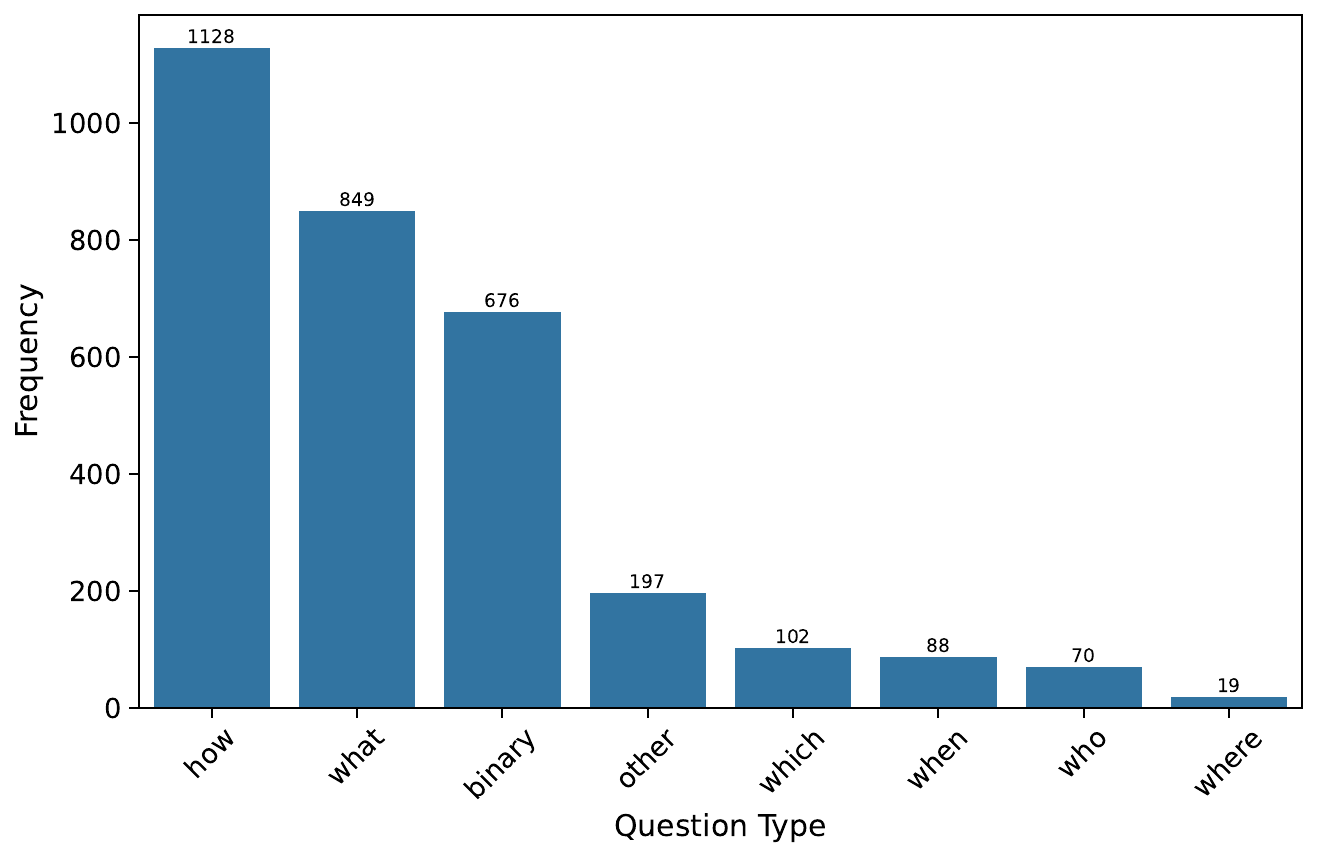}
    \caption{Distribution of types of questions in VLQA.}
    \label{fig:dist_type_ques}
\end{figure}

To investigate the diversity of VLQA, we examine the question from three perspectives: question types, reasoning types, and question topics. First, we manually classified questions into 8 types: How, What, Binary, Which, When, Who, Where, Other, as shown in Table \ref{tab:describe_type_ques}. Figure \ref{fig:dist_type_ques} shows that the ``How'' question accounts for the highest proportion at 36\%, followed by the ``What'' question at 27.1\%. The types ``Which'', ``When'', ``Who'', and ``Where'' contribute a small proportion (together less than 9\%). 


\begin{table}[t]
\caption{Descriptions and examples of different types of questions in VLQA.}
\label{tab:describe_type_ques}
\begin{tabular}{l|p{5.8cm}|p{5.8cm}}
\hline
\multicolumn{1}{c|}{\textbf{Question Type}} & \multicolumn{1}{c|}{\textbf{Description}}                                                                           & \multicolumn{1}{c}{\textbf{Examples}}                                                                                                                                         \\ \hline
How                                        & Question concern about the method of doing legal matters & How are means of transporting cash, valuable assets and valuable papers regulated? \\ 
What                                       & Question focuses on definitions, objects, concepts                                                                 & What are the responsibilities of the driver and service staff on the bus?                           \\ 
Binary                                     & Question has a Yes/No or True/False answer option.                                                             & Is an unnotarized civil contract valid?                                                                             \\ 
Which                                      & Question involves choices, options, where the answer is one or more options presented within the legal ground. & In which cases must animal products be quarantined at the destination?                                        \\
When                                       & Question regards a specific amount of time.                                                                    & When is the deadline for returning the results of the business registration certificate?        \\ 
Who                                        & Question identifies an organization or individual.                                                             & Who has the authority to issue a certificate of operation for a driving test center?    \\ 
Where                                      & Question related to a location or position.                                                                    & Where should a business with many branches file a bankruptcy petition?                      \\ 
Other                                      & Question concerns about how many, how much, or content that are not categorized in the above types                 & How much is the fine for installing a power source for a karaoke bar without permission?                  \\ \hline
\end{tabular}
\end{table}

Table \ref{tab:describe_reasoning_type} categorizes legal reasoning into four distinct types: word matching, concept understanding, logic, and multi-article reading. Among these, semantic interpretation accounts for the majority of instances (59\%), highlighting the prevalence of questions that require an understanding of legal semantics, synonyms, antonyms, and deeper interpretations. 
In contrast, the lexical matching category is characterized by a high lexical similarity without external context. While this type facilitates easier retrieval, it represents the most surface-level form of reasoning. Questions in this group often rely on direct phrase alignment rather than contextual understanding, making them more tractable for retrieval-augmented models or those relying on keyword matching.
The logical inference category, which comprises approximately one-fifth of the data, involves binary reasoning such as Yes/No questions or conditional statements. This type of reasoning requires models not only to identify relevant legal content but also to apply logical operations or deductions based on that content.
Finally, the multi-article reading type, accounting for about 25\% of the dataset, represents the most cognitively demanding form of reasoning. These questions require integrating and reconciling information across multiple legal provisions or articles, often involving cross-referencing, and temporal reasoning (i.e., repealed law). 


\begin{table}[]
\caption{Distribution and examples of different reasoning types of questions in VLQA. One question may require multiple reasoning abilities, so the sum of percentages is over 100\%.}
\label{tab:describe_reasoning_type}
\begin{tabular}{l|c|p{9cm}}
\hline
\multicolumn{1}{c|}{\textbf{Reasoning Type}} & \textbf{Percentage}  & \multicolumn{1}{c}{\textbf{Examples}}                                                                                                                                                                                                                   \\ \hline
Lexical Matching                                & 41\%                  & \textbf{Question}: How are working relationships regulated in handling incidents of the national power system?                                                             \\
                                             &                       & \textbf{Answer}: Working relationships in handling incidents of the national power system are regulated in [...]                                                \\
                                             &                       & \textbf{Article 14}. Working relationships in handling incidents                                                                                                                                          \\ \hline
Semantic Interpretation                       & 59\%                  & \textbf{Question}: How many hours does each regular vocational training session last?                                                                                                                                                                            \\
                                             &                       & \textbf{Answer}: When you participate in a regular vocational training program, the maximum study time per session is 05 (five) hours and the maximum study time in a day is 08 (eight) hours                                                                    \\
                                             &                       & \textbf{Article 9}. Training time and plan 1. Training time for regular training programs specified in [...]                                                                                                                                                \\ \hline
Logical Inference                                        & 22\%                  & \textbf{Question}: Do I need to make a customs declaration when exporting software via electronic means?                                                                                                                                                         \\
                                             &                       & \textbf{Answer}: If your establishment conducts software export business via electronic means, it does not need to make a customs declaration.                                                                                                                   \\
                                             &                       & \textbf{Article 16}. Conditions for deduction and refund of input tax on exported goods and services                                                                                                                                                             \\ \hline
Multi-Article Reading                        & 25\%                  & \textbf{Question}: If the employee does not take time off work when adopting a child, how much will the one-time social insurance benefit be?                                                                                                                    \\
                                             &                       & \textbf{Answer}: In case the employee adopts a child who is eligible for social insurance leave but does not take time off, he/she will receive a one-time benefit for each child equal to 02 times the basic salary at that month \\
                                             & \multicolumn{1}{l|}{} & \textbf{Article 31}. Procedures for implementing the regime for employees who are not eligible for pension and have not reached the age of receiving social pension benefits                                                                                     \\
                                             & \multicolumn{1}{l|}{} & \textbf{Article 38}. Rate of compulsory social insurance and voluntary social insurance contributions                                                                                                                                                            \\ \hline
\end{tabular}
\end{table}

\begin{figure}
    \centering
    \includegraphics[width=\linewidth]{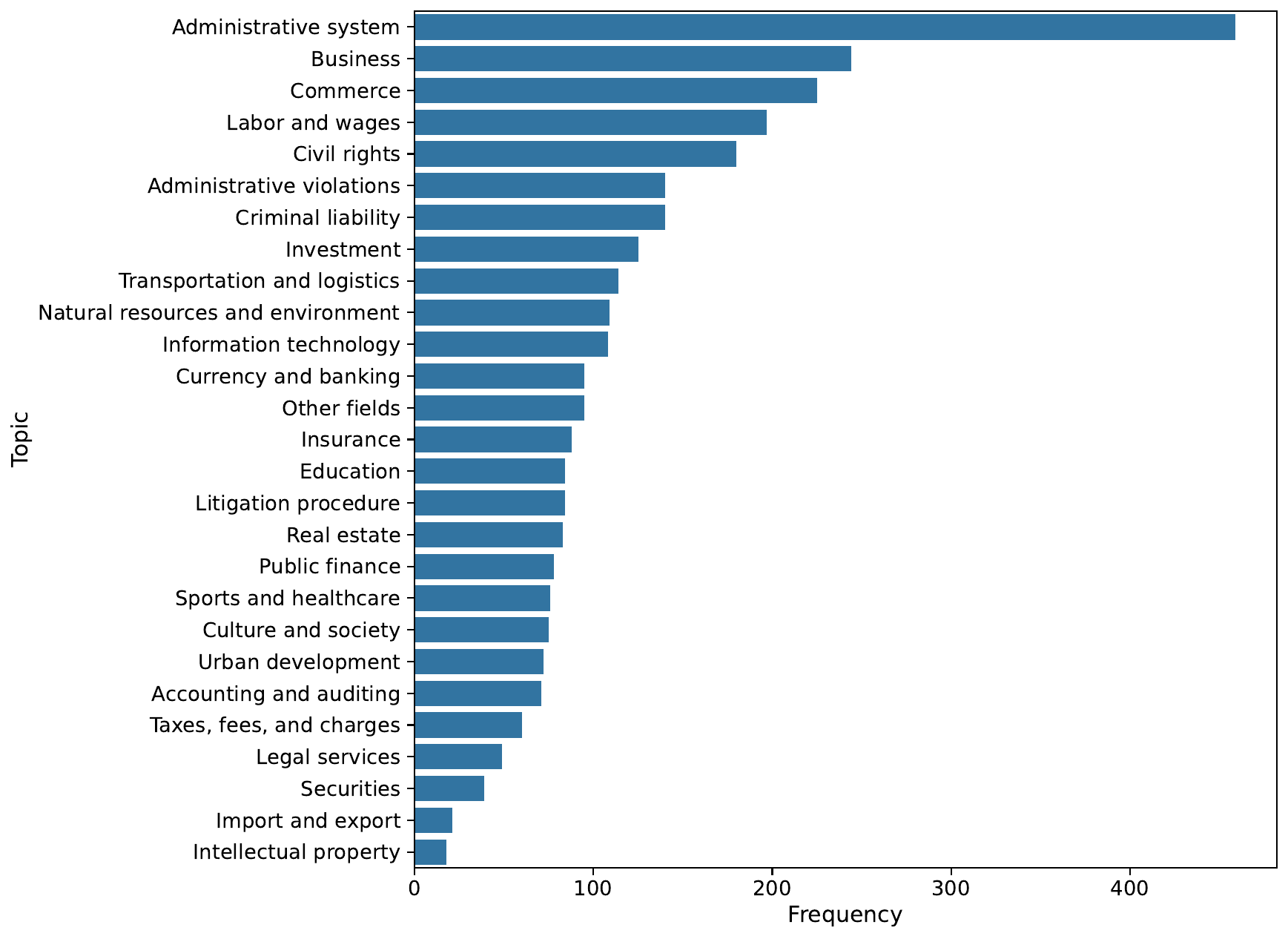}
    \caption{Distribution of questions by law category in VLQA.}
    \label{fig:dist_ques_topic}
\end{figure}

To assess the topic coverage of the VLQA dataset, we analyze the distribution of legal questions across different subject areas, as illustrated in Figure \ref{fig:dist_ques_topic}. Topics are assigned based on the relevant legal articles associated with each question. The Administrative System domain accounts for the largest share, comprising 14.67\% of all questions, approximately twice the proportion of questions in the Business category. In contrast, questions related to Intellectual Property, Import and Export, and Securities represent the smallest proportions, each contributing less than 3\% of the total dataset. This distribution reflects the varying degrees of public engagement with different areas of law.


\subsubsection{Question-article reference}
Questions in the VLQA dataset may be associated with one or multiple relevant legal articles. As shown in Figure \ref{fig:no_rel_law}, 94\% of the questions reference at most two articles, while the remaining 6\% cite between three and nine. These multi-article references often involve more complex legal reasoning or require supplemental information to address repealed or cross-referenced provisions, which explains the higher number of associated articles.


\section{Models}
\label{sec:models}
We now describe the baseline models, ranging from traditional methods to advanced models for legal article retrieval (Section \ref{sec:ldr}) and question-answering tasks (Section \ref{sec:lqa}). 

\subsection{Task 1: Legal Article Retrieval}
\label{sec:ldr}

Mathematically, given a question $q$ and a legal corpus $\mathbf{A}$, the retrieval system $\mathcal{R}:(q,\mathbf{A}) \to \mathcal{F}$ is a function that outputs top$-k$ candidate articles $\mathbf{C}=\{c_i\}^k_{i=1} \subset \mathbf{A}$ that is relevant to the question $q$. These articles are ranked by order of relevance and then selected by different strategies. The following section presents retrieval models used as benchmarks for this task. 

\subsubsection{Sparse Retrieval Models: }
Traditionally, sparse retrieval models have been essential for information retrieval due to their efficiency in handling large-scale data. The score for a sequence can be computed by summing or averaging the weights of its words. Following the same mechanism, BM25 \cite{bm25} is a probabilistic model based on the occurrence frequency of query terms in documents. The relevance score of a query $q$ and legal article $a$ is calculated as follows:
\begin{equation}
    \operatorname{BM25}(q, a)=\sum_{i=1}^{|q|} \frac{\operatorname{IDF}\left(t_i\right) \cdot T F\left(t_i, d\right) \cdot\left(k_1+1\right)}{T F\left(t_i, a\right)+k_1 \cdot\left(1-b+b \cdot \frac{|a|}{\text { avgdl }}\right)}
    \label{eq:bm25}
\end{equation}
where $TF(t_i, a)$ presents the term frequency of $t_i$  in the article $a$, $IDF(t_i)$ represents inverse document frequency, $|q|$ and $|a|$ is the length of query $q$ and article $a$, respectively, $avgdl$ is the average length of all articles in the corpus. $k_1 \in \mathbb{R}_{+}, b \in [0,1]$ are free hyperparameters that could be tuned on the development set. We compute the relevance score of each question to all articles in the corpus using BM25 and return top$-k$ most relevant articles to the query $q$. 

Another sparse retrieval approach is applying Doc2Vec \cite{pmlr-v32-le14} to produce vector representations for documents. Following this method, we generate sparse vector representations for query $q$ and article $a$ using fastText \cite{bojanowski2017enriching}. A similarity scoring function is then applied to compute the relevance score of a question-answer pair. Similarly to BM25, we also return top$-k$ the most relevant articles to the query $q$.


\subsubsection{Dense Retrieval Models}
\paragraph{Zero-shot Retrieval Approach}
Despite their promises, sparse retrieval models have shown limited success in information retrieval \cite{10.1145/3308774.3308781,10.1145/3331184.3331340}.
\citet{reimers-gurevych-2019-sentence} proposed Sentence-BERT, a siamese architecture for deriving semantically meaningful sentence embeddings. This bi-encoder network has been applied successfully in various tasks such as textual entailment, question answering, and information retrieval. Therefore, we apply the bi-encoder architecture to perform legal article retrieval, where the question and article are encoded independently into dense vectors. The similarity score between these vectors computes the relevance of the question-article pair. Mathematically, the bi-encoder method is formulated as follows:

\begin{equation}
     \text{Relevance}(q,a) = f(\phi(q), \psi(a))
\end{equation}
where $\phi$ and $\psi$ are functions, or encoders, i.e, BERT \cite{devlin-etal-2019-bert}, BGE \cite{chen-etal-2024-m3}, mapping question $q$ and article $a$ to their representation spaces with dimensional size $d$, i.e., 768. Function $f:\mathbb{R}^d \times \mathbb{R}^d \rightarrow \mathbb{R}$ indicates a similarity function such as dot product or cosine score. 

\paragraph{Supervised learning Approach:}

\begin{figure}[t]
\centering
\includegraphics[width=0.8\linewidth,keepaspectratio]{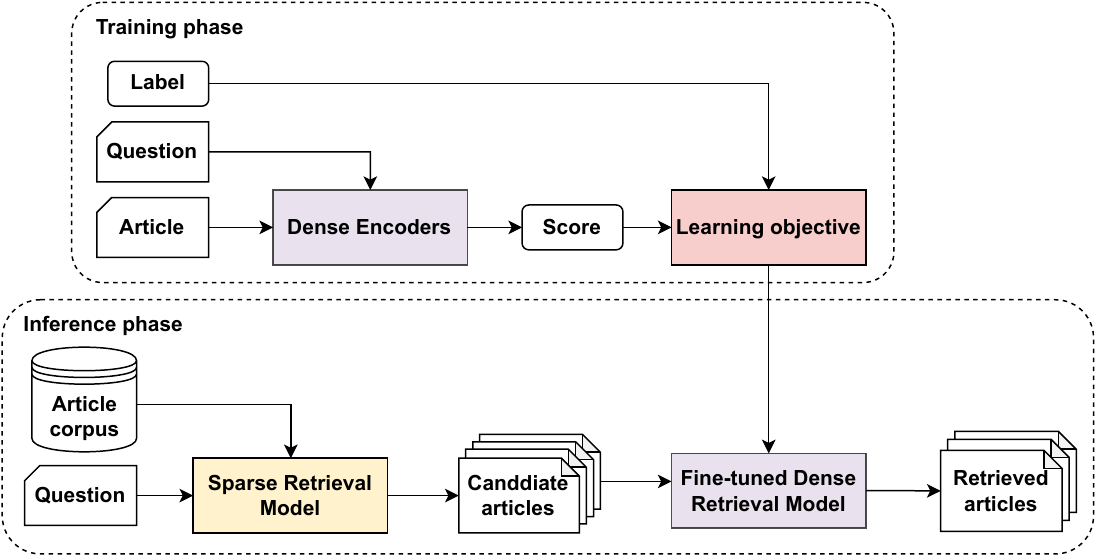}
\caption{Supervised learning framework for the legal article retrieval task in VLQA.} 
\label{fig:retrieval_pipeline}
\Description[Retrieval pipeline]{Illustration of the legal article retrieval task performed on VLQA.}
\end{figure}

Thereafter, we introduce a two-phase framework based on a lexical retrieval model and a fine-tuned cross-encoder architecture, as shown in Figure \ref{fig:retrieval_pipeline}. The lexical model serves as a pre-ranking step for the latter model. Let the training set be $\mathcal{T}= \{ \langle q, a, y \rangle_i \}_{i=1}^N$, where article $a \in \{ \mathbf{A}^+ \cup \mathbf{A}^-\}$ with $\mathbf{A}^+$ is the set of relevant (positive) articles and  $\mathbf{A}^-$ indicates the set of negative articles derived from top$-k$ outputs of the previous phase, $y\in \{0,1\}$ is the label, and $N$ presents number of samples in the training set. Since following the cross-encoder architecture, we optimize the retriever with a binary cross-entropy loss, as below:
\begin{equation}
        \mathcal{L}_{bce}{(q,a,y)} = -\frac{1}{N} \sum_{i=1}^N y_i \log(\phi(q,a)_i) + (1 - y_i)\log(1-\phi(q,a)_i)
\end{equation}
where $\phi(q,a)$ is an encoder mapping a pair of $q$ and $a$ to the dimensional space, $y_i$ takes the value 1 if article $a_i$ should be labeled as a relevant article, otherwise 0. 

\subsection{Legal Question Answering}
\label{sec:lqa}

For the legal question answering task, the input contains a pair of legal question $q$ and top$-k$ candidate articles $\mathbf{A} = \{a_i\}^k_{i=1}$. The system aims to generate a span of text $\mathbf{S}=\{s_1, s_2, \dots, s_n\}$ that can address or answer the question. We tackle this task using two different approaches: extractive and generative models. 
Extractive models cut short phrases from legal articles to serve as answers, preventing models from generating nonsense and new words. Meanwhile, generative models learn a predictor function $f(q,\mathbf{A}) \rightarrow \mathbf{S}$ to generate a short and concise answer, which can be formulated as a conditional text generation problem. In this section, we investigate two learning scenarios: supervised training models and the in-context learning ability of LLMs \cite{radford2019language}. 

\subsubsection{Supervised learning QA models}
\begin{figure}[t]
\centering
\includegraphics[width=0.5\linewidth,height=\textheight,keepaspectratio]{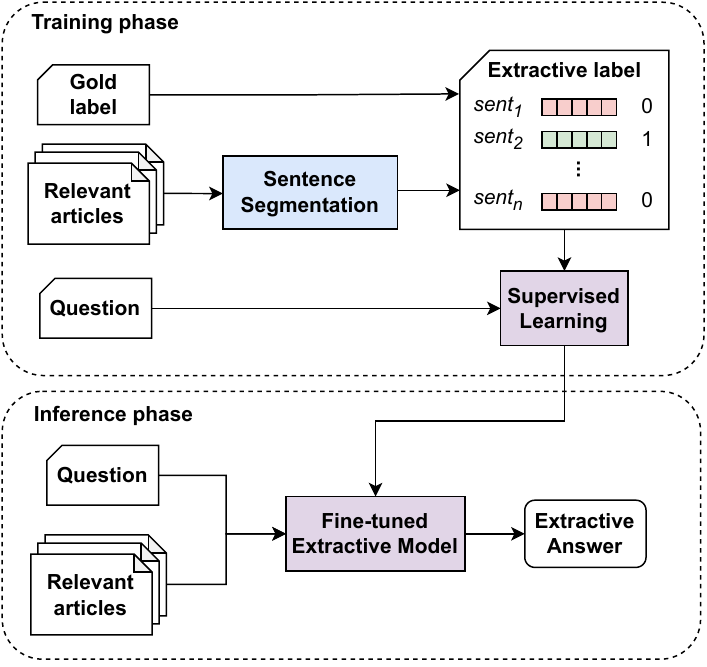}
\caption{An extractive QA framework for the legal question answering task in VLQA.} 
\label{fig:ex_pipeline}
\Description[Extractive pipeline]{Illustration of extractive approach for question answering task performed on VLQA.}
\end{figure}

\paragraph{Fine-tuned Extractive Models: } 
Extractive QA models identify specific words and phrases in legal texts to generate answers. Therefore, these models could not be trained directly on the gold-label dataset created by legal experts. The primary reason is that the answers in the gold-label dataset may not directly map to the exact wording or phrasing in the legal articles. To address this challenge, we construct a silver-label dataset. This dataset will consist of new answers, denoted as $\mathbf{S}^\prime$, which are extracted directly from the legal articles instead of human-generated answers in the gold-label dataset. Figure \ref{fig:ex_pipeline} presents the construction steps of the silver-label dataset and training extractive models. Legal articles are segmented into sentences, denoted as $\mathbf{a}^\prime=\{a'_i\}^L_{i=1}$ using Python-based tools. Each sentence is then mapped to the gold-label answer to determine which sentence in legal articles is the answer. Finally, the selected sentences are combined to form a silver answer. Different variants of BERT are employed in this approach. The training set of BERT-based models is a set of triplets $\mathcal{T} = \{ \langle q, a', y \rangle_i \}_{i=1}^N$ where $y\in\{0,1\}$ is the label, $N$ is the number of samples. The output is the probability of a sentence being included in the answer, computed as follows:
\begin{equation}
    \textbf{H}_{cls} = \phi(q,a_i)
\end{equation}
\begin{equation}
    \textbf{P}_{label=1}(q,a) = softmax(\textbf{H}_{cls} * W_f)
\end{equation}
where $H_{cls} \in \mathbb{R}^d$ is the final vector presentation of the pair $(q, a_l)$ produced by a cross-encoder model $\phi$, $W_f \in \mathbb{R}^{d \times 2}$ is the weight matrix of the final fully connected layer, and $softmax$ is an activation function. 
The model is trained using the cross-entropy loss between the model's logits and target. 
\begin{equation}
    \mathcal{L}_{ce} (q,a,y) = - \sum_{c=1}^2 \log  \frac{\text{exp}({s(q, a)_c})}{\sum_{j=1}^2 \text{exp}({s(q,a)_j})} y
\end{equation}

\paragraph{Fine-tuned Generative Models: }
For the generative approach, we utilize two recent Vietnamese pre-trained models ViT5 \cite{phan2022vit5} and BARTpho \cite{bartpho}. ViT5 is an encoder-decoder model for the Vietnamese language based on T5-style self-supervised learning \cite{raffel2020exploring}. BARTPho is a pre-trained sequence-to-sequence model for Vietnamese based on the pretraining scheme of the denoising autoencoder BART \cite{lewis-etal-2020-bart}. Both models have the same input size, which is 1024 tokens. Formally, given a training sample $(q, \mathbf{A}, \mathbf{S}$) where $\mathbf{S} = \{s_1, s_2, \dots, s_n\}$ is the answer or target output, we optimized parameters of generative models using a cross-entropy loss function of generating the target answer conditioned on the input context. 
\begin{equation}
    \begin{split}
        \mathcal{L}(q, \mathbf{A},\mathbf{S})  & =  - \log \textbf{P}(\mathbf{S} \, | \, q \, , \mathbf{A}  ) \\
        & = - \log \prod_{t=1}^n \textbf{P} (s_t \, | \, q \, , \mathbf{A} \, , s_{<t})
    \end{split}
\end{equation}



\subsubsection{In-context learning} 
LLMs have achieved state-of-the-art performance across various NLP downstream tasks, demonstrating strong capabilities in acquiring general knowledge and following task-specific instructions \cite{10.1145/3641289}. In this study, we evaluate the performance of several recent LLMs, including Qwen2.5, GPT-4o, and DeepSeek-V3, via the in-context learning method. Specifically, we investigate two widely adopted prompting settings: zero-shot and few-shot learning.
\paragraph{Zero-shot prompting}
LLMs learn to answer the legal question directly based on the input articles via their in-context learning capability. The input template $input$ provided to the model is formulated as 
\begin{equation}
    input = [inst, \mathbf{A}, q]
\end{equation}
where $inst$ meaning provided instruction, $q$ is the legal question and $\mathbf{A}$ is the set of top$-k$ relevant articles determined by the retriever. 

\paragraph{Few-shot prompting}
In few-shot learning, models are provided with ground-truth examples drawn from the development set to help guide their understanding of the task. These examples, which serve as demonstrations for the response format and correct reasoning paths, can be included within the model’s context window. Thus, the input template for few-shot can be defined as follows:
\begin{equation}
    input = [inst, [q_i, \textbf{A}_i, \textbf{S}_i]_{i=1}^m, \textbf{A}, q]
\end{equation}
where $\textbf{S}_i$ is the annotated answer of the example question $q_i$, and $m$ indicates the number of examples for few-shot, and here we set $m=2$.

\section{Experiments}
\label{sec:experiment}
We now benchmark state-of-the-art methods on the VLQA test set, focusing on three key aspects: \textit{retrieval performance}, \textit{answer quality}, and \textit{factual accuracy}. We employ both automatic metrics and human evaluations to provide a comprehensive assessment of the baseline models' performance across these dimensions.


\subsection{Evaluation Measures} 
\label{sec:eval_methods}

\subsubsection{Automatic evaluation:}
We examine the performance of retrieval models on six automatic metrics, including recall$@k$, precision$@k$, F2$@k$, mean average precision (MAP$@k$), and mean reciprocal rank (MRR$@k$),  normalized discounted cumulative gain (NDCG$@k$). 
The objective of legal document retrievers is to retrieve as many relevant articles as possible within the top-$k$ results. To evaluate retrieval effectiveness, we primarily use the recall score, defined as the ratio of correctly retrieved articles to the total number of relevant articles for a given question. However, recall alone may not fully capture retrieval quality, as increasing $k$ inherently boosts recall.
To address this limitation, we also report the precision score, which measures the proportion of correctly retrieved articles among all retrieved articles. Finally, we include the F-score, the harmonic mean of precision and recall, to provide a balanced evaluation of retrieval performance. The F-score is calculated as follows:
\begin{equation}
    {F}_\beta = (1 + \beta^2) \times \frac{\text{precision} \times \text{recall}}{(\beta^2 \times \text{precision}) + recall}
\end{equation}
In this work, we utilize the F2 score, which means $\beta=2$, assigning weights to recall higher than precision.

To evaluate our question‐answering system automatically, we employ both ROUGE \cite{lin2004rouge} and BERTScore \cite{zhang2020bertscore}. ROUGE measures the overlap of $n$-grams between the generated and reference texts; in this study, we report ROUGE-1, ROUGE-2, and ROUGE-L to capture unigram recall, bigram recall, and longest‐common‐subsequence performance, respectively. While these lexical overlap metrics are widely used in summarization, they often underperform on open‐ended QA tasks in the legal domain, where semantically equivalent but lexically divergent spans abound \cite{krishna-etal-2021-hurdles,xu-etal-2023-critical}. To address this, we also report BERTScore, which computes the similarity between candidate and reference sentences in a high-dimensional embedding space of a pre-trained transformer. Prior work has demonstrated that BERTScore correlates more closely with human judgments at the sentence level, making it especially suitable for assessing the nuanced, context‐sensitive answers typical of legal QA \cite{zhang2020bertscore}.

\paragraph{Human evaluation:}
Evaluating legal question-answering systems is inherently multifaceted, demanding not only automatic metrics but also expert judgment to assess factual accuracy and answer completeness. Accordingly, beyond reporting ROUGE and BERTScore, we engaged qualified legal professionals to review model outputs, flagging errors of fact, omission, and misinterpretation. Their assessments enabled us to categorize common failure modes, such as misapplication of legal terms and redundant pieces of information, to derive meaningful insights into the strengths and limitations of LLMs in legal QA.

\subsection{Experimental Setup} 
\label{sec:exp_setup}

This section presents detailed experimental configurations in two tasks: legal article retrieval and question answering. Experiments are conducted on an NVIDIA A100 GPU with 40GB VRAM, using PyTorch library. 

\subsubsection{Retrieval models}
The recall score is reported at $k = [1,2,5,10,20,50,100,200,500,1000]$ to provide a comprehensive analysis of the baseline. Other retrieval metrics are computed at $k=2$. Hyperparameters of BM25 are set as $k_1=1.2,b=0.75$.
For the zero-shot retrieval approach, we apply pre-trained checkpoints of a Vietnamese SBERT\footnote{\url{https://huggingface.co/keepitreal/vietnamese-sbert}} and BGE-m3\footnote{\url{https://huggingface.co/BAAI/bge-m3}}. For the supervised learning approach, mBERT\footnote{\url{google-bert/bert-base-multilingual-cased}} and BGE-reranker\footnote{\url{https://huggingface.co/BAAI/bge-reranker-v2-m3}} are optimized for 5 epochs using the Adam optimizer \cite{kingma2017adam} with an initial learning rate of  $2e^{-5}$, batch size of 32, number of negative articles $|\mathbf{A}^-|=5$. The maximum sequence length of mBERT and BGE-rerank is 512 and 1024 tokens, respectively. 
The article pool of supervised retrievers is reduced from all articles to the top $200$ articles derived from BM25.
The evaluation metrics used to assess the retrieval performance are Recall$@k$, Precision$@k$, F2$@k$, MAP$@k$, MRR$@k$, and NDCG$@k$. 


\subsubsection{Question-answering models}
Evaluation metrics for this task are ROUGE-1, ROUGE2, ROUGE-L \cite{lin2004rouge}, BERTScore \cite{zhang2020bertscore}, and expert evaluation. For extractive QA models, we fine-tune the following models: mBERT-base~\cite{devlin-etal-2019-bert}, LEGAL-BERT-base\cite{chalkidis-etal-2020-legal}, RoBERTa-base~\cite{conneau-etal-2020-unsupervised}, PhoBERT-base~\cite{nguyen-tuan-nguyen-2020-phobert}, and ViDeBERTa-base~\cite{tran-etal-2023-videberta} on the silver training set for 5 epochs, using the Adam optimizer, initial learning rate of $2e^{-5}$, batch size of 32. The maximum sequence length of these models is 512 tokens. Therefore, we perform splitting for sequences longer than the context window into multiple input features. 

For generative QA models, we employ ViT5-base and BARTpho-word-base for sequence-to-sequence QA systems. These models are fine-tuned for 5 epochs, using Adam optimizer, with an initial learning rate of $2e^{-5}$, batch size of 4, and maximum output tokens of 1024. Both models have a maximum input sequence of 1024 tokens. Before fine-tuning BARTpho, we perform word segmentation using VnCoreNLP \cite{vu2018vncorenlp}. During the inference phase, we use the greedy decoding setting for these generative models to ensure consistency for later evaluation. 

For the in-context learning setup, we evaluate a diverse set of both open-weight and commercial large language models. Our open-weight experiments include Qwen2.5-7B-Instruct and Qwen2.5-14B-Instruct, each supporting a context window of 32,768 tokens, four times that of the Llama series, such as Meta Llama3-8B-Instruct and Llama3.1-8B-Instruct. All open-weight models are hosted locally using the vLLM framework for high-throughput inference \cite{kwon2023efficient}. In parallel, we benchmark commercial LLMs (GPT-4o, GPT-4o-mini, and DeepSeek-V3) via their respective APIs. 
We also apply a greedy decoding setting to ensure consistency for later evaluation. The instruction prompts for in-context learning are shown in Appendix \ref{sec:appendix_prompt}. The zero-shot instruction for LLMs is constructed similarly, without containing example information.



\section{Results}
\label{sec:results}
We now report the detailed results of state-of-the-art retrieval and legal question-answering methods on the VLQA test set. In addition to quantitative evaluation using established automatic metrics, we perform a meticulous manual analysis to gain deeper insights into the qualitative aspects of LLMs' performance, particularly their reasoning ability, and linguistic coherence in legal contexts.

\subsection{Legal Article Retrieval}
\label{sec:exp_lar}

Table~\ref{tab:retrieval_results_unsup} presents the recall performance of both sparse and dense retrieval models under zero-shot settings. Among all models, BGE-m3 demonstrates a clear advantage, significantly outperforming other baselines. This performance likely stems from its use of advanced training methodologies and access to large-scale training data. BM25 continues to serve as a robust baseline for domain-specific retrieval tasks, offering competitive recall scores alongside high computational efficiency. In contrast, bi-encoder models using fastText and SBERT show poor results when applied without task-specific fine-tuning. This observation is consistent with prior findings by \citet{reimers-gurevych-2019-sentence}, emphasizing the limitations of using pre-trained embeddings in domain-specific retrieval tasks without adaptation. 
These findings underscore the potential of recent instruction-tuned dense retrieval models like BGE-m3 in addressing real-world legal tasks. 
We present a comprehensive evaluation of diverse retrieval methods under both zero-shot and supervised settings, focusing on top-$2$ returned articles, as illustrated in Figure~\ref{tab:f2_retrieval}. Our results show that the fine-tuned mBERT dense retriever, trained on 2.1k in-domain VLQA samples, substantially outperforms strong zero-shot baselines, underscoring the critical importance of domain adaptation for effective legal information retrieval. However, all models show precision scores smaller than 0.40, indicating that less than half of the retrieved articles are relevant, on average. These results suggest a substantial room for enhancing the performance of retrievers in terms of accuracy. 

\begin{table}[t]
    \centering
    \caption{Recall scores of retrieval approaches on the test set in the zero-shot setting.}
    \label{tab:retrieval_results_unsup}
\begin{tabular}{l|cccccccccc}
\hline
\textbf{Model} & \textbf{R@1} & \textbf{R@2} & \textbf{R@5} & \textbf{R@10} & \textbf{R@20} & \textbf{R@50} & \textbf{R@100} & \textbf{R@200} & \textbf{R@500} & \textbf{R@1000} \\ \hline
fastText       & 0.1380       & 0.2281       & 0.3310       & 0.4044        & 0.4781        & 0.5446        & 0.6006         & 0.6713         & 0.7475         & 0.7986          \\
SBERT          & 0.1274       & 0.1733       & 0.2841       & 0.3579        & 0.4476        & 0.5695        & 0.6518         & 0.7325         & 0.8100         & 0.8522          \\
BM25           & 0.2670       & 0.3835       & 0.5221       & 0.6139        & 0.6928        & 0.7678        & 0.8232         & 0.8651         & 0.9113         & 0.9318          \\
BGE-m3         & 0.4323       & 0.5442       & 0.6928       & 0.7814        & 0.8449        & 0.9099        & 0.9348         & 0.9610         & 0.9691         & 0.9800          \\ \hline
\end{tabular}
\end{table}


\begin{table}[t]
    \centering
    \caption{Retrieval results of unsupervised and supervised models on the test set.}
    \label{tab:f2_retrieval}
\begin{tabular}{lcccccc}
\hline
\multicolumn{1}{l|}{\textbf{Model}} & \textbf{Precision@2} & \textbf{Recall@2} & \textbf{F2@2}   & \textbf{MAP@2}  & \textbf{MRR@2}  & \textbf{NDCG@2} \\ \hline
\multicolumn{7}{l}{\textbf{One-stage retrievers}}                                                                                                      \\ \hline
\multicolumn{1}{l|}{SBERT}          & 0.1045               & 0.1733            & 0.1531          & 0.1808          & 0.1794          & 0.1883          \\
\multicolumn{1}{l|}{fastText}       & 0.1348               & 0.2281            & 0.2003          & 0.2139          & 0.2153          & 0.2313          \\
\multicolumn{1}{l|}{BM25}           & 0.2265               & 0.3835            & 0.3368          & 0.3278          & 0.3780          & 0.4017          \\
\multicolumn{1}{l|}{BGE-m3}         & 0.3222               & 0.5442            & 0.4783          & 0.4921          & 0.5654          & 0.5925          \\ \hline
\multicolumn{7}{l}{\textbf{Two-stage retrievers}}                                                                                                      \\ \hline
\multicolumn{1}{l|}{BGE-rerank}     & 0.3437               & 0.5845            & 0.5127          & 0.5504          & 0.6467          & 0.5869          \\
\multicolumn{1}{l|}{mBERT-ft}       & \textbf{0.3740}       & \textbf{0.6257}   & \textbf{0.5515} & \textbf{0.5757} & \textbf{0.6675} & \textbf{0.6179} \\ \hline
\end{tabular}

\end{table}

\subsection{Legal Question Answering}
\label{sec:exp_lqa}

Table~\ref{tab:qa_gen_results} presents the evaluation results of both extractive and generative models on the VLQA test set using automated metrics, including ROUGE and BERTScore. As discussed in Section~\ref{sec:lqa}, we assess the models’ ability to generate accurate and contextually appropriate legal answers.
Overall, extractive models tend to outperform generative models on lexical overlap metrics (ROUGE-1 and ROUGE-2), which measure word-level similarity with reference answers, indicating their strong capability in capturing surface-level content overlap. However, generative models excel in contextual semantic similarity, as shown by higher BERTScore. 
Notably, LEGAL-BERT, despite being pre-trained on legal-domain data and fine-tuned on the VLQA training set, performs the worst across all metrics. This suggests the importance of language understanding in addressing domain-specific downstream tasks. 

\begin{table}[ht]
\caption{Evaluation results of extractive and generative models on VLQA legal question answering.}
\label{tab:qa_gen_results}
\begin{tabular}{lcccc}
\hline
\multicolumn{1}{l|}{\textbf{Model}} & \multicolumn{1}{l}{\textbf{ROUGE-1}} & \multicolumn{1}{l}{\textbf{ROUGE-2}} & \multicolumn{1}{l}{\textbf{ROUGE-L}} & \multicolumn{1}{l}{\textbf{BERTScore}} \\ \hline
\multicolumn{5}{l}{\textbf{Extractive models}}                                                                                                                                                 \\ \hline
\multicolumn{1}{l|}{LEGAL-BERT}     & 0.4528                               & 0.4235                               & 0.2271                               & 0.6549                              \\
\multicolumn{1}{l|}{ViDeBERTa}      & 0.5065                               & 0.4754                               & 0.2530                               & 0.7104                              \\
\multicolumn{1}{l|}{mBERT}          & 0.5114                               & 0.4786                               & 0.2561                               & 0.7090                              \\
\multicolumn{1}{l|}{XLM-RoBERTa}    & 0.5187                               & 0.4872                               & 0.2617                               & 0.7314                              \\
\multicolumn{1}{l|}{PhoBERT}        & \textbf{0.5423}                      & \textbf{0.5088}                      & 0.2683                               & 0.7304                              \\ \hline
\multicolumn{5}{l}{\textbf{Generative models}}                                                                                                                                                 \\ \hline
\multicolumn{1}{l|}{ViT5}           & 0.4332                               & 0.3198                               & 0.3210                               & 0.7666                              \\
\multicolumn{1}{l|}{BARTpho}        & 0.4695                               & 0.3827                               & \textbf{0.3318}                      & \textbf{0.7872}                     \\ \hline
\end{tabular}
\end{table}

Table~\ref{tab:LLMs_results} presents the performance of recent open-weight and commercial LLMs on the VLQA legal question-answering task under both zero-shot and two-shot prompt settings. Among all models evaluated, GPT-4o-mini consistently achieves the highest scores across all metrics and settings, demonstrating its strong generalization capabilities in complex legal QA tasks. Notably, Qwen2.5-14B stands out among open-weight models, highlighting the strength of the Qwen family in handling Vietnamese legal text. An important observation is that two-shot prompting yields mixed results: while it enhances performance in larger models such as GPT-4o and DeepSeek-V3, it can degrade performance in smaller LLMs (i.e., models with fewer than 14 billion parameters). This decline may stem from input length limitations and the reduced intrinsic memory capacity of smaller models, which struggle to effectively leverage few-shot examples. Finally, it is noteworthy that fine-tuned BARTpho, despite being substantially smaller, achieves performance comparable to Llama3.1-8B, bolstering the effectiveness of domain adaptation in addressing legal downstream tasks. 

\begin{table}[t]
\caption{Performances of recent open-weight and commercial LLMs on VLQA legal question answering.}
\label{tab:LLMs_results}
\begin{tabular}{l|clclclcl}
\hline
\textbf{Model}       & \multicolumn{2}{c}{\textbf{ROUGE-1}}                  & \multicolumn{2}{c}{\textbf{ROUGE-2}}                  & \multicolumn{2}{c}{\textbf{ROUGE-L}}                  & \multicolumn{2}{c}{\textbf{BERTScore}}                   \\
                     & \textbf{0-shot} & \multicolumn{1}{c}{\textbf{2-shot}} & \textbf{0-shot} & \multicolumn{1}{c}{\textbf{2-shot}} & \textbf{0-shot} & \multicolumn{1}{c}{\textbf{2-shot}} & \textbf{0-shot} & \multicolumn{1}{c}{\textbf{2-shot}} \\ \hline
Llama3.1-8B-Instruct & 0.4384          & 0.4313                              & 0.3709          & 0.3681                              & 0.3807          & 0.3779                              & 0.7431          & 0.7414                              \\
Qwen2.5-7B-Instruct  & 0.4988          & 0.4854                              & 0.4190          & 0.4102                              & 0.4250          & 0.4153                              & 0.7263          & 0.7231                              \\
Llama3-8B-Instruct   & 0.5720          & 0.5698                              & 0.4967          & 0.4966                              & 0.5007          & 0.5005                              & 0.7745          & 0.7739                              \\
Qwen2.5-14B-Insutrct & 0.6017          & 0.5995                              & 0.5138          & 0.5118                              & 0.5197          & 0.5179                              & 0.7614          & 0.7613                              \\
DeepSeek-V3          & 0.6704          & 0.6809                              & 0.5307          & 0.5631                              & 0.5165          & 0.5567                              & 0.7769          & 0.8022                              \\
GPT-4o               & 0.6824          & 0.7040                               & 0.5917          & 0.5993                              & 0.5979          & 0.5912                              & 0.8296          & 0.8233                              \\
GPT-4o-mini          & \textbf{0.6975} & \textbf{0.7137}                     & \textbf{0.6011} & \textbf{0.6219}                     & \textbf{0.6001} & \textbf{0.6192}                     & \textbf{0.8314} & \textbf{0.8339}                     \\ \hline
\end{tabular}
\end{table}

Table~\ref{tab:llms_ques_type_res} presents a detailed evaluation of recent LLMs across various question types under a two-shot prompting setting, aiming to assess their legal reasoning capabilities. Overall, both Qwen2.5-14B-Instruct and GPT-4o-mini achieve the highest scores on lexical matching questions, which primarily require surface-level similarity detection. However, their performance progressively declines on more reasoning-intensive tasks such as semantic interpretation and logical inference, highlighting limitations in handling complex domain-specific questions. These findings underscore the need for further research into domain adaptation techniques and the enhancement of reasoning capabilities in LLMs for legal applications. 

\begin{table}[t]
\caption{Performance of recent LLMs on different types of questions under two-shot prompting setting. }
\label{tab:llms_ques_type_res}
\resizebox{\columnwidth}{!}{
\begin{tabular}{l|cccc|cccc}
\hline
\textbf{Reasoning Type} & \multicolumn{4}{c|}{\textbf{Qwen2.5-14B-Instruct}}                          & \multicolumn{4}{c}{\textbf{GPT-4o-mini}}                                    \\
                        & \textbf{ROUGE-1} & \textbf{ROUGE-2} & \textbf{ROUGE-L} & \textbf{BERTScore} & \textbf{ROUGE-1} & \textbf{ROUGE-2} & \textbf{ROUGE-L} & \textbf{BERTScore} \\ \hline
Lexical Matching        & 0.6520           & 0.5796           & 0.5811           & 0.7826             & 0.7709           & 0.7007           & 0.6983           & 0.8592             \\
Multi-Article Reading   & 0.6211           & 0.5210           & 0.5165           & 0.7754             & 0.6857           & 0.5744           & 0.5650           & 0.8154             \\
Semantic Interpretation & 0.5579           & 0.4581           & 0.4679           & 0.7445             & 0.6684           & 0.5595           & 0.5566           & 0.8138             \\
Logical Inference                   & 0.4835           & 0.3605           & 0.3772           & 0.7242             & 0.5619           & 0.4389           & 0.4460           & 0.7882             \\ \hline
All questions                   & 0.5995           & 0.5118           & 0.5179           & 0.7613             & 0.7137           & 0.6219           & 0.6192           & 0.8339             \\ \hline\end{tabular}}
\end{table}

\subsection{Qualitative Analysis}

\begin{table}[t]
\caption{Qualitative analysis of LLMs' responses on 100 randomly selected samples from the test set.}
\label{tab:qualitivative_analysis}
\begin{tabular}{lp{5cm}ccc}
\hline
\textbf{Error Type} & \textbf{Description}                 & \textbf{PhoBERT} & \textbf{Qwen2.5-14B} & \textbf{GPT-4o-mini} \\ \hline
\textbf{Syntactic errors}    & Errors in answer's format, languages & 12    & 40          & 0           \\
\textbf{Semantical errors}   & Redundant answer                  & 11             & 2                    & 2                    \\
                    & Incomplete answer                 & 22             & 8                    & 6                    \\
                    & Logical incorrect                    & 0              & 2                    & 5           \\  \hline
\textbf{Total \# errors}      &                                      & 45             & 52                   & 13                   \\ \hline
\end{tabular}
\end{table}

To better understand the strengths and limitations of QA methods, including an extractive model and LLMs, we conduct a detailed manual analysis of 100 randomly selected samples from the test set, as shown in Table~\ref{tab:qualitivative_analysis}. Each output is carefully evaluated across two key dimensions: syntactic errors, referring to formatting issues or the presence of non-Unicode characters; and semantic errors, where the answer is fluent and natural but logically incorrect or lacks critical information.
Overall, answers generated by LLMs tend to be shorter than the gold answers. Specifically, the average lengths of responses from Qwen2.5 and GPT-4o-mini are 159 and 172 words, respectively, compared to an average of 198 words for gold answers. This discrepancy helps explain why many of the semantic errors observed in LLM outputs come from incomplete answers. 
Regarding the performance of GPT-4o-mini, it demonstrates notable strengths in producing coherent, human-like answers. However, most of the generated outputs are incomplete or logically incorrect, particularly in handling logical inference questions. We also observe that GPT-4o-mini tends to replicate entire article texts instead of effectively addressing the given legal question.
Another noteworthy finding is that, despite achieving stronger performance on automatic metrics, Qwen2.5-14B shows lower accuracy than PhoBERT in human evaluations. Among its incorrect outputs, 77\% come from syntactic errors, including nonsensical phrasing or non-Unicode characters. This highlights the limitations of smaller LLMs in handling complex, long-form legal QA tasks.
Finally, while LLMs can generate detailed answers based on provided facts and conditions, they might contain inaccurate or hallucinated elements, ultimately leading to unreliable conclusions, as reported in Appendix \ref{sec:appendix_llm_hallu}.

\section{Conclusions}
\label{sec:conclusion}
In this work, we introduce VLQA, a comprehensive and high-quality dataset designed for the Vietnamese legal domain. As part of a broader effort to bridge the gap between the public and legal knowledge, VLQA provides a solid foundation for the development and evaluation of reliable models in legal downstream tasks. We benchmark a range of state-of-the-art retrieval and question-answering approaches, evaluated through both automated metrics and human review, to demonstrate the dataset's utility and highlight areas for future improvement. Looking ahead, we aim to develop a robust end-to-end framework capable of handling lengthy statutory texts, complex legal queries, and real-world application needs. We hope that VLQA will inspire further research into practical and trustworthy legal question-answering systems, ultimately contributing to equal access to the law for all.

\begin{acks}
We would like to express our appreciation to the data collection team from the University of Engineering and Technology, Vietnam National University, Hanoi. We also extend our heartfelt thanks to the data annotation team from the University of Law, Vietnam National University, Hanoi, for their valuable contribution to this work. 
\end{acks}

\bibliographystyle{ACM-Reference-Format}
\bibliography{sample-base,custom}

\appendix
\section{In-context learning instruction for LLMs in legal question-answering}
\label{sec:appendix_prompt}
The instructions for zero-shot and two-shot prompting are defined as follows: 
\begin{lstlisting}[caption={Zero-shot instruction learning for LLMs in legal question answering.}, captionpos=t]
You are a legal expert specializing in Vietnamese law, with in-depth knowledge of legal regulations and their practical applications. Your task is to answer legal questions accurately, clearly, and professionally.

### Instructions:
1. It is mandatory to provide answers in Vietnamese.
2. Base your answers solely on the provided information and avoid adding any assumptions or external knowledge.
3. Ensure that your response is well-structured, concise, and relevant to the question.
4. If the provided information does not contain the answer, state that clearly and suggest seeking further clarification.

Here is the question and the relevant information to answer that question

### Question:
{question}

### Relevant Information:
{context}
\end{lstlisting}

\begin{lstlisting}[caption={Few-shot instruction learning for LLMs in legal question answering.}, captionpos=t]
You are a legal expert specializing in Vietnamese law, with in-depth knowledge of legal regulations and their practical applications. Your task is to answer legal questions accurately, clearly, and professionally.

### Instructions:
1. It is mandatory to provide answers in Vietnamese.
2. Base your answers solely on the provided information and avoid adding any assumptions or external knowledge.
3. Ensure that your response is well-structured, concise, and relevant to the question.
4. If the provided information does not contain the answer, state that clearly and suggest seeking further clarification.

Here are examples of how to answer questions based on the provided information:

Example 1:
Question: {question_1}
Relevant law articles: {context_1}
Answer: {answer_1}

Example 2:
Question: {question_2}
Relevant law articles: {context_2}
Answer: {answer_2}

Here is the question and the relevant information to answer that question

### Question:
{question}

### Relevant Information:
{context}
\end{lstlisting}

\section{Illustrations of LLMs' incorrect responses}
\label{sec:appendix_llm_hallu}
We provide detailed examples of semantic errors and hallucinated outputs produced by LLMs, as identified through qualitative analysis. Table~\ref{tab:sem_error_llm} highlights specific mistakes made by GPT-4o-mini and Qwen-2.5-14B regarding the monetary fine for driving a motorbike against traffic flow. These errors are likely attributable to the models’ insufficient domain knowledge, particularly their inability to accurately interpret the relevant legal scenario and extract relevant information from the provided legal articles. Furthermore, we identify a hallucination pattern in GPT-4o-mini, where it fabricates information not present in the source text, such as inventing the name of a legal document, illustrated in Table~\ref{tab:llm_fabric_lawname}.


\begin{table}[ht]
\caption{Semantic error of LLMs due to extracting an incorrect piece of information within the provided context.}
\label{tab:sem_error_llm}
\begin{tabular}{lp{11.5cm}}
\hline
\textbf{Question}    & What is the fine for driving a motorbike against the traffic?                                                                                                        \\  \hline
\textbf{Gold answer} & Motorcyclists driving in the opposite direction will be fined from \textbf{400,000 VND to 600,000 VND}.                                                                       \\ \hline
\textbf{GPT-4o-mini}           & Motorcyclists driving in the opposite direction will be fined from \textbf{1,000,000 VND to 2,000,000 VND} according to the provisions of Article 6 of Decree 100/2019/ND-CP.  \\ \hline
\textbf{Qwen-2.5-14B} & According to Decree 100/2019/ND-CP, driving a motorbike against traffic will be fined from \textbf{2,000,000 VND to 3,000,000 VND}. Specifically, Article 5, Clause 5 of Decree 100/2019/ND-CP stipulates: ``A fine of 2,000,000 VND to 3,000,000 VND shall be imposed on a driver who commits one of the following violations: d) Driving against traffic on a one-way street, driving against traffic on a road with a ``no going against traffic'' sign, except for violations specified in Point b, Clause 6 of this Article and cases where priority vehicles are on urgent missions as prescribed''. Therefore, if you are caught riding a motorbike against traffic, you will be fined from \textbf{2,000,000 VND to 3,000,000 VND}. \\ \hline

\end{tabular}
\end{table}

\begin{table}[]
\caption{Hallucination behavior of GPT-4o-mini in fabricating name of law.}
\label{tab:llm_fabric_lawname}
\begin{tabular}{lp{11cm}}
\hline
\textbf{Question}              & What are the principles for determining tuition fees for preschools?                                                                                                                             \\ \hline
\textbf{Gold answer}           & Pursuant to Clause 1, \textbf{Article 8 of Decree 81/2021/ND-CP}, the principles for determining tuition fees for preschool education institutions are specified as follows: Tuition fees are established according to the principle of sharing between the Government and learners, in accordance with the socio-economic conditions of each residential area, the actual contribution capacity of the people and the growth rate of the consumer price index, the annual economic growth rate, the roadmap for calculating prices of educational and training services according to regulations and ensuring the quality of education. With this provision, the principles for determining tuition fees for preschool education institutions will be established according to the above provisions.                                                                                                                                                                                                                                                                            \\ \hline
\textbf{GPT-4o-mini}           & Pursuant to \textbf{Article 8 of the 2019 Law on Education}, the principles for determining tuition fees for preschool education institutions are stipulated as follows: 1. Tuition fees are established based on the principle of sharing between the Government and learners, in accordance with the socio-economic conditions of each residential area, the actual contribution capacity of the people and the growth rate of the consumer price index, the annual economic growth rate, the roadmap for calculating prices of educational and training services according to regulations and ensuring the quality of education. Thus, tuition fees at preschool education institutions will be determined based on socio-economic factors and the contribution capacity of the people.                                                                                                                                                                                                                     \\ \hline
\textbf{Qwen-2.5-14B-Instruct} & According to \textbf{Article 8 of the provided legal document}, the principle of determining tuition fees for preschool education institutions is stipulated as follows: Tuition fees are established according to the principle of sharing between the Government and learners, in accordance with the socio-economic conditions of each residential area, the actual contribution capacity of the people and the rate of increase in the consumer price index, the annual economic growth rate, the roadmap for calculating prices of educational and training services according to regulations and ensuring the quality of education. Specifically, determining tuition fees must consider many factors such as the socio-economic conditions of the locality, the contribution capacity of parents, the rate of increase in the consumer price index and annual economic growth, as well as the roadmap for calculating prices of educational and training services. At the same time, tuition fees also need to ensure the quality of education provided by the preschool. \\ \hline 
\end{tabular}
\end{table}

\end{document}